\documentclass{article}
\usepackage{iclr2026_conference,times}

\usepackage{amsmath,amsfonts,bm}

\def\eqref#1{equation~\ref{#1}}

\def\1{\bm{1}}

\DeclareMathAlphabet{\mathsfit}{\encodingdefault}{\sfdefault}{m}{sl}
\SetMathAlphabet{\mathsfit}{bold}{\encodingdefault}{\sfdefault}{bx}{n}

\usepackage{hyperref}
\usepackage{url}
\usepackage{booktabs}
\usepackage{multirow}
\usepackage{amsmath,amssymb}
\usepackage{graphicx}
\usepackage{xcolor}
\usepackage{microtype}

\newcommand{\bench}{\textsc{DevDataBench}}

\title{Field Order Should Not Matter: Permutation-Invariant Embedding Model Fine-Tuning for\\Structured Metadata Retrieval}

\author{Aivin V. Solatorio \\
Development Data Group \\
Office of the Chief Statistician, World Bank Group \\
\texttt{asolatorio@worldbank.org}
\And
Olivier Dupriez \\
Development Data Group \\
Office of the Chief Statistician, World Bank Group \\
\texttt{odupriez@worldbank.org}
\And
Rafael S. Macalaba \\
Development Data Group \\
Office of the Chief Statistician, World Bank Group \\
\texttt{rmacalaba@worldbank.org}}

\iclrfinalcopy
\begin{document}

\maketitle

\begin{abstract}
We study retrieval over catalogs of structured metadata, where each record is a small schema whose fields answer different kinds of query. Embedding a record with a text encoder first serializes its fields into a string, which forces a choice of field order. We show this choice, usually treated as an implementation detail, silently controls retrieval quality once the encoder is fine-tuned. A standard fine-tune loses 7.4 nDCG@10 points when the index is rebuilt under a different field order, because it reads absolute position instead of the field labels. We propose \emph{permutation-invariant fine-tuning} (\textbf{PI-FT}), which serializes each record under a freshly sampled field order with random field dropout, so meaning binds to the labels rather than to position. The change is about two lines in the data loader; it costs negligible in-distribution accuracy and cuts the order-change penalty to 0.2 points. We study this in the discovery of development statistics, a catalog of nearly 10{,}000 indicators that should be searchable in many languages by a model small enough to self-host. As AI assistants and agents increasingly mediate access to public data and statistics, this retrieval step decides whether an answer is grounded in the right indicator or series, making discoverability a precondition for disseminating data through AI. Because usage logs cannot provide training signal for indicators no one has searched, we generate the queries instead. \bench{} is a fully LLM-generated benchmark of grounded, facet-targeted queries across 15 languages, covering every indicator for both training and evaluation. A fine-tuned 118M-parameter encoder, runnable on a CPU, outperforms every zero-shot baseline including \texttt{text-embedding-3-large} (0.707 vs.\ 0.556 nDCG@10), with the largest gains on the low-resource languages those users speak. We release the benchmark, pipeline, and models, and the training framework applies to any labeled-field catalog.
\end{abstract}

\section{Introduction}
\label{sec:intro}

At the core of data-driven policymaking are social and economic indicators whose value depends not only on their quality, but also on how easily they can be discovered and accessed. Public data portals catalog thousands of such indicators, and which one a query retrieves is decided entirely by the indicator's metadata. That decision is increasingly made by the retrieval layer beneath an AI agent rather than by a person scrolling a portal, which raises the cost of getting it wrong. Consider a query for ``domestic value added in exports for Japan in 2015.'' The terms that identify the right indicator are scattered across a structured record. The concept sits in the name and definition. The country and year are constraints the record must satisfy through its geographic-coverage list and time-coverage window. Many queries depend on fields the name never states, such as how a series was measured, in what unit, or by which agency. This is retrieval over structured records, where each document is a schema and a query can target any of its fields. To embed a record with a text encoder, we first flatten its fields into a string, and that forces a choice of order.

That order is normally treated as an implementation detail, but it changes results. Once an encoder is fine-tuned on one serialization, the field order it saw in training effectively becomes part of the model. In our experiments, a standard fine-tuned encoder loses 7.4 nDCG@10 points when the index is rebuilt under a different field order, and even a frontier zero-shot embedding (\texttt{text-embedding-3-large}) loses 6.4. The cause is structural. A transformer has no built-in notion that a record means the same thing whatever order its fields appear in, so it is free to read meaning off absolute position. The same spurious order sensitivity has been documented for tabular encoders \citep{yang-etal-2022-tableformer}. Any catalog that re-renders its metadata after a portal redesign, or that ingests records from producers who serialize differently, will quietly degrade a model trained this way.

The fix needs no architectural change and works with any text encoder. We call it \emph{permutation-invariant fine-tuning} (\textbf{PI-FT}). We serialize each training record as labeled field segments, using the catalog's own field names as the labels, so the framework needs no hand-curated schema. At every step we present the record under a freshly sampled field order and drop non-essential fields at random. When the same record keeps reappearing under different orders, position stops being a useful cue, and the only stable anchor left is the field label. The encoder is pushed to read the schema rather than memorize a template. The change is about two lines in the data loader. It costs negligible accuracy on the model's own serialization and cuts the order-change penalty from 7.4 points to 0.2.

The augmentation is not new in spirit. It builds on a long line of work on permutation-invariant representation learning (Section~\ref{sec:related}), and field-order permutation has already been used to serialize tabular rows for \emph{generative} modeling \citep{borisov2023great, solatorio2023realtabformer}. What is new is where we study it, the contrastive fine-tuning of retrieval embeddings over serialized statistical metadata. To our knowledge, we are the first to identify and remove serialization-order fragility in the contrastive fine-tuning of retrieval embeddings over structured metadata. We establish three things. First, fine-tuned metadata retrievers and zero-shot embeddings are severely order-fragile, with nDCG@10 dropping by 9 to 26\% when the index field order changes. Second, permutation augmentation carries over from generation to retrieval and removes this fragility at negligible in-distribution cost. Third, we give the framework a theoretical account, as a Janossy-style approximation to group averaging over the field-permutation group (Section~\ref{sec:theory}).

We study this in the discovery of development statistics, a setting where the engineering point has real stakes. Development data is curated mostly in English, but the analysts, journalists, researchers, and AI systems that use it work in many languages, and the national statistical offices, international organizations, and ministries that steward these catalogs \citep{worldbank2021wdr} often cannot route metadata or queries through external services and cannot absorb a per-query API charge. The setting therefore calls for an encoder that is schema-aware, multilingual, and small enough to self-host. It is also a demanding test of order-invariance. These catalogs re-render their metadata after portal redesigns and ingest records from producers who serialize differently, so a retriever that reads field position rather than field labels degrades exactly when the catalog changes. Findability is the first of the FAIR principles \citep{wilkinson2016fair}, and a series that cannot be found is collected again or ignored, so the gains here translate into data that is actually used.

Search logs, even where they exist, would supervise only the indicators users have already found, since click feedback is biased toward what was shown and selected \citep{joachims2017unbiased}, leaving the long tail of never-searched series with no signal \citep{park2008longtail}. This is a cold-start problem \citep{schein2002coldstart} that log mining cannot solve. We therefore generate the training and evaluation queries ourselves, covering every indicator, language, and facet uniformly, including questions no user has issued yet. Generating retrieval supervision from documents is by now well established and competitive with human-labeled data \citep{nogueira2019doc2query,bonifacio2022inpars,dai2023promptagator,wang-etal-2022-gpl}. We contribute:

\begin{itemize}
\item \textbf{Order fragility in metadata retrieval, and permutation-invariant fine-tuning (PI-FT) as a remedy} (Sections~\ref{sec:theory}--\ref{sec:method}): we quantify how strongly embedding retrievers depend on arbitrary field order, provide a group-theoretic justification for permutation-invariant fine-tuning, and show how it combines naturally with full-schema serialization, facet-protected dropout, similarity-filtered hard negatives, and guided contrastive learning. Component ablations attribute the gains to each ingredient, while a two-condition robustness evaluation (canonical versus permuted indexes) isolates invariance from retrieval quality across encoder sizes.

\item \textbf{\bench{}}, a benchmark for structured-metadata retrieval over the World Bank Data360 catalog (9{,}948 indicators, 161 databases): 765k facet-targeted synthetic training queries and 32{,}442 held-out evaluation queries generated across thirteen search facets and 15 languages, with every query grounded in record content. The benchmark incorporates leakage controls (disjoint indicator splits and a different generator family for evaluation), together with a quantified near-duplicate audit tailored to statistical catalogs. Every artifact can be regenerated directly from the public catalog at low cost (Appendix~\ref{app:cost}).

\item \textbf{Evidence that supervision quality can outweigh parameter count for specialized retrieval}: a 118M-parameter encoder trained with our framework outperforms every zero-shot baseline, including open models up to 600M parameters and frontier API embeddings, while remaining CPU-deployable. A systematic comparison across small open encoders, together with a tokenizer audit, identifies where the framework delivers the largest gains and where tokenizer vocabulary becomes the limiting factor rather than the training method \citep{petrov2023tokenizers}. Finally, we show that guided contrastive training compounds through self-distillation: using a first-generation model trained with the same framework as the GIST guide yields a further improvement without sacrificing permutation invariance.
\end{itemize}

\begin{figure}[t]
\centering
\includegraphics[width=\linewidth]{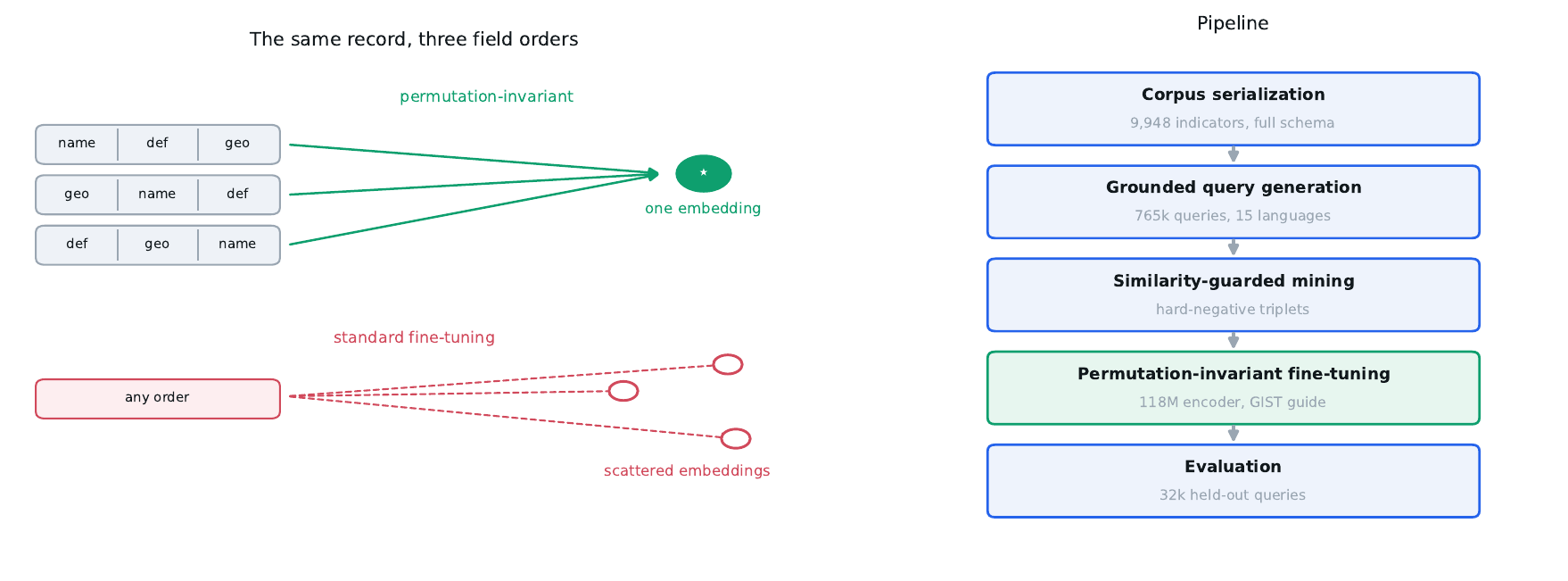}
\caption{Overview. Left: the same record serialized under three field orders collapses to one embedding under permutation-invariant fine-tuning (PI-FT), but scatters under standard fine-tuning. Right: the pipeline, from corpus serialization through grounded multilingual query generation, similarity-guarded hard-negative mining, and permutation-invariant (GIST-guided) fine-tuning to evaluation, with data volumes.}
\label{fig:overview}
\end{figure}

\section{Related Work}
\label{sec:related}

\paragraph{Text embeddings and retrieval benchmarks.}
Contrastively trained bi-encoders are the standard architecture for dense retrieval \citep{reimers-gurevych-2019-sentence,karpukhin-etal-2020-dense,gao-etal-2021-simcse,izacard2022contriever}, scaled through multi-stage frameworks \citep{wang2022e5,li2023gte,xiao2023bge} and trained largely on web and QA corpora such as MS MARCO \citep{bajaj2016msmarco}. Evaluation is anchored by BEIR \citep{thakur2021beir} and MTEB \citep{muennighoff-etal-2023-mteb}, and on the multilingual axis by mMARCO \citep{bonifacio2021mmarco} and MIRACL \citep{zhang2023miracl}. None of these covers retrieval over statistical metadata, where queries target schema fields and the corpus is a catalog of structured records rather than passages.

\paragraph{Multilingual representation learning.}
Massively multilingual encoders \citep{devlin2019bert,conneau2020xlmr,feng2022labse} trade per-language capacity for coverage, the ``curse of multilinguality'' \citep{conneau2020xlmr}. Multilingual E5 \citep{wang2024me5} and related models bring this lineage to retrieval. Our setting is cross-lingual by construction. The documents stay in English while queries arrive in 15 languages, which is the realistic configuration for international statistical catalogs and the one multilingual encoders are pretrained to support. We further characterize how \emph{monolingual}-vocabulary encoders fail under multilingual fine-tuning at the tokenizer level, which complements work on tokenizer-induced inequity across languages \citep{petrov2023tokenizers}.

\paragraph{Synthetic supervision for retrieval.}
Generating queries from documents is an established substitute for missing logs, as in doc2query \citep{nogueira2019doc2query}, InPars \citep{bonifacio2022inpars}, Promptagator \citep{dai2023promptagator}, and GPL \citep{wang-etal-2022-gpl}. These works show generated queries can train retrievers that match supervised baselines from a handful of demonstrations \citep{dai2023promptagator}, and can adapt a retriever to a new domain with no labels at all \citep{wang-etal-2022-gpl}, which is the closest prior analog to our cold-start setting. Generation also covers what a log cannot: the long tail of indicators no one has searched, and the full facet and language grid, for both training and evaluation. Our generation differs in being schema-targeted, grounded, and multilingual. Facets are assigned per record from its populated fields, and constraint facets must use values drawn from the record itself. We add two controls that prior synthetic-retrieval work omits. The held-out set is produced by a different model family than the training set, and identifier-lookup queries are generated deterministically rather than paid for once per language.

\paragraph{Negative selection and false negatives.}
Hard negatives drive contrastive quality \citep{xiong2021ance}, but harder mining also surfaces more \emph{false} negatives. Denoising them with a cross-encoder \citep{qu2021rocketqa} or a guide embedding model \citep{solatorio2024gist} recovers the lost signal. In catalog retrieval this denoising is structurally necessary, because statistical corpora are saturated with near-duplicates (Section~\ref{sec:corpus}) and dense synthetic supervision makes same-positive batch collisions frequent. Section~\ref{sec:gist} quantifies both effects.

\paragraph{LLMs as annotators and judges.}
LLM relevance judgments approach human agreement at a fraction of the cost \citep{faggioli2023llmjudge,thomas2024llmrel,zheng2023judging}. We also use LLMs as the complete annotation layer, covering training queries, evaluation queries, and graded relevance. In place of the guarantees that human annotation would provide, we rely on explicit audits, namely generator separation, near-duplicate quantification, and grounding checks.

\paragraph{Permutation invariance and order sensitivity.}
Whether a model's output should be invariant to the ordering of set-structured inputs is a long-studied question. Deep Sets \citep{zaheer2017deepsets} characterizes permutation-invariant functions, Janossy pooling \citep{murphy2019janossy} builds them by averaging a permutation-sensitive function over input orderings, with single-permutation stochastic training as the tractable approximation, and Set Transformer \citep{lee2019settransformer} realizes the property through attention. The group-theoretic analysis of data augmentation \citep{chen2020grouptheoretic} shows that augmenting over a group's orbit is equivalent to projecting onto the invariant subspace. This is the formal basis for inducing invariance through data rather than through architecture. Transformers, in the opposite direction, are known to be spuriously \emph{order-sensitive}, both to context position \citep{liu2024lostmiddle} and to the ordering of in-context examples \citep{lu2022fantastically}. Our setting exhibits both effects. Serialized records are order-sensitive, which we measure, and field-order permutation is the orbit-averaging fix. The same permutation operation has been used to serialize tabular rows for generative LLMs \citep{borisov2023great}, and we transfer it to contrastive retrieval fine-tuning and analyze its effect there.

\paragraph{Dataset search and structured retrieval.}
Dataset search engines \citep{brickley2019datasetsearch,chapman2020dataset} and table retrieval \citep{zhang2020webtable} establish that metadata carries retrieval signal of its own. Production systems handle structure with faceted interfaces \citep{tunkelang2009faceted}, query parsing \citep{guo2009nerq}, field-weighted lexical scoring \citep{robertson2004bm25f}, and hybrid fusion \citep{cormack2009rrf}. These compose with a schema-aware encoder rather than competing with it (Section~\ref{sec:discussion}). TableFormer \citep{yang-etal-2022-tableformer} achieves order robustness architecturally for tables, whereas we achieve it for serialized records purely through data augmentation, which applies to any encoder. Closest in goal is SANTA \citep{li2023santa}, which improves dense retrieval over structured data through structure-aware \emph{pretraining}, using structured-data alignment and entity masking. We instead target the serialization-order dependence introduced at fine-tuning time, a problem that is orthogonal to and compatible with structure-aware pretraining.

\section{Problem Setting and Corpus}
\label{sec:corpus}

\begin{figure}[t]
\centering
\includegraphics[width=\linewidth]{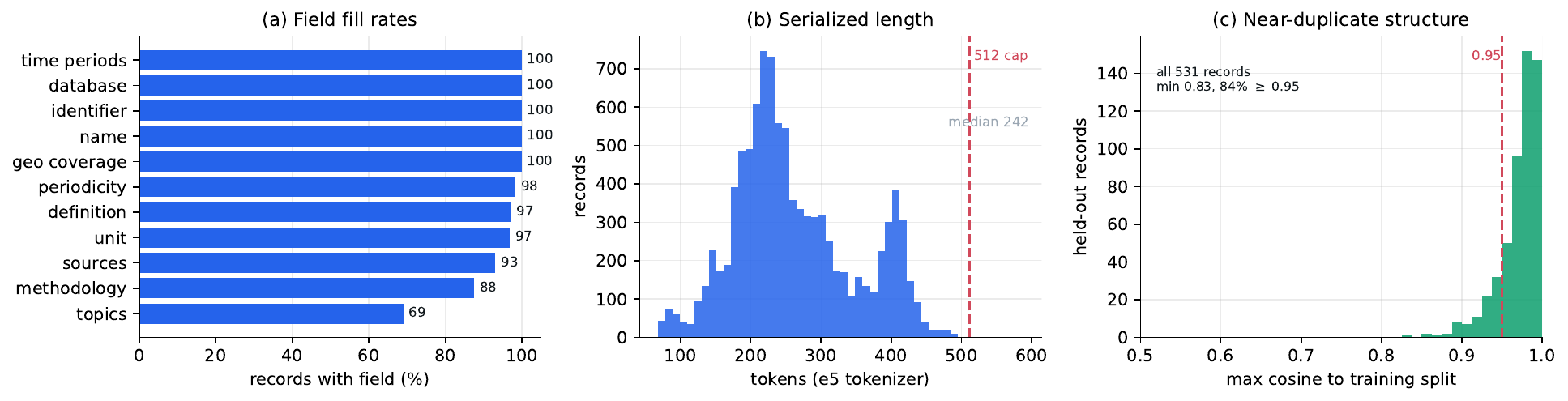}
\caption{Corpus anatomy. (a) Field fill rates across the 9{,}948 records. (b) Token-length distribution of full-schema serializations (e5 tokenizer) with the 512-token cap marked. (c) Each held-out record's maximum cosine similarity to the training split (NoInstruct base encoder), with the 0.95 near-duplicate threshold marked. All 531 held-out records are shown; the minimum is 0.83, so every one has a near neighbor in training. This raises a leakage concern that Section~\ref{sec:results-neardup} tests directly: models score at least as well on the records with no close training neighbor, so the near-duplication does not inflate the results.}
\label{fig:corpus}
\end{figure}

The corpus is the indicator catalog of the World Bank Data360 platform, which holds 9{,}948 metadata records from 161 source databases (World Development Indicators, IMF, ITU, UNESCO, OECD, FAO, among others). Each record is a structured schema of labeled fields. All of them carry a name, identifier code, database affiliation, geographic-coverage list, and time-period list. Beyond that, 98.3\% carry periodicity, 97.2\% a definition, 96.9\% a measurement unit, 93.1\% source organizations, 87.5\% methodology text, and 69.1\% a topic taxonomy (Figure~\ref{fig:corpus}a). Serialized in full (Section~\ref{sec:serialization}), records have a median length of 242 tokens, a 90th percentile of 402, and a maximum of 588, under the XLM-RoBERTa tokenizer of the \texttt{multilingual-e5} base (209, 344, and 542 under a BERT WordPiece tokenizer).

Two structural properties shape everything downstream. The first is that \textbf{queries distribute over the schema}. A definition question resolves against the definition field, ``ITU connectivity statistics'' against the source, and ``donn\'ees annuelles depuis 2002'' against periodicity and time coverage. A representation that drops a field cannot answer queries about it. The second is that \textbf{the catalog is saturated with near-duplicates}. International organizations republish one another's series, so the same conceptual indicator appears under several identifiers. When we embed all records with a base encoder, the median held-out record has a maximum cosine similarity of \emph{0.977} to some training record, and 84.2\% exceed 0.95. This property alone drives our split audit (Section~\ref{sec:splits}), the hard-negative similarity guard, and the case for guided training (Section~\ref{sec:gist}).

\section{\bench{}: A Benchmark for Structured-Metadata Retrieval}
\label{sec:bench}

\subsection{Facet-targeted, grounded query generation}
\label{sec:generation}

For each indicator, an LLM receives the record's full serialization and a sampled set of \emph{facets}, and must write queries that the record answers. Twelve content facets cover the schema: \texttt{keyword}, \texttt{natural}, \texttt{definition}, \texttt{methodology}, \texttt{geo}, \texttt{year}, \texttt{geo\_year}, \texttt{unit}, \texttt{source}, \texttt{database}, \texttt{frequency}, and \texttt{thematic}. Facet assignment follows field availability, so a record with no methodology text gets no methodology facet. Constraint facets are grounded in the record. A \texttt{geo} query must name an economy on the coverage list or a region that contains one, and a \texttt{year} query must fall inside a real coverage window. A thirteenth facet, \texttt{code} (identifier lookup), is generated \emph{deterministically} from templates over the identifier. These queries are exact and language-neutral, so sending them through an LLM would only multiply the cost by the number of languages.

\paragraph{Languages.}
Queries are generated in 15 languages spanning seven scripts. These are English, Spanish, French, Portuguese, German, Indonesian, Swahili, Turkish, Russian, Arabic, Hindi, Bengali, Urdu, Japanese, and Chinese, chosen to cover the working languages of major development institutions and the home languages of large developing-country populations. Documents remain in English, so the task is cross-lingual retrieval, which matches the operational reality of international catalogs. Coverage is \emph{full} rather than sampled, meaning every (indicator, language) pair receives queries and every language sees the entire catalog. Full coverage is the reason to generate rather than log: it exercises every field in every language, including questions no user has issued yet, so a model trained on it is not bounded by historically popular queries. To keep the cost of full coverage linear in queries rather than in records times languages, languages are grouped five per request. The metadata is sent once and queries for all five languages are returned with per-item language tags, which cuts input cost by roughly $5\times$. Generation used a small model (\texttt{claude-haiku-4-5}) through a 50\%-discounted batch API, and codes, acronyms, and proper nouns are kept untranslated by instruction.

The training split comprises about 765k queries over 9{,}417 indicators across 15 languages, plus 18{,}834 templated code queries (two per indicator). Failure handling is conservative: malformed responses are dropped (a small fraction of the 28{,}251 generation requests) and exact duplicate (record, query) pairs are removed.

\begin{figure}[t]
\centering
\includegraphics[width=\linewidth]{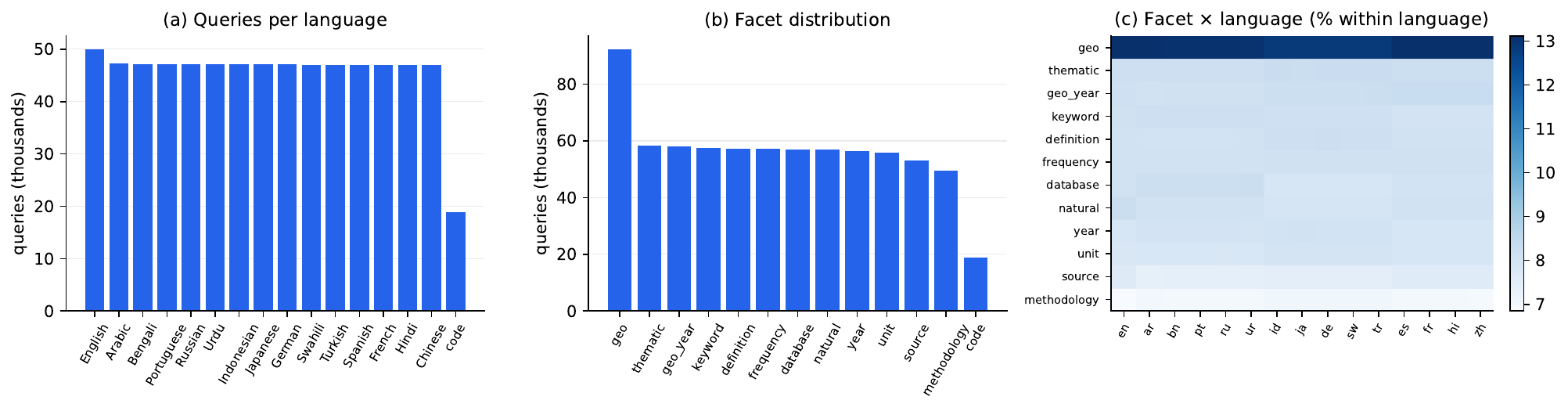}
\caption{Training-data anatomy. (a) Queries per language (15 languages plus templated code queries, near-uniform by design). (b) Facet distribution. (c) Facet $\times$ language as a percentage within each language, showing dense coverage of the design grid.}
\label{fig:traindata}
\end{figure}

\subsection{Splits, leakage controls, and the near-duplicate audit}
\label{sec:splits}

Indicators are assigned to train (95\%) or evaluation (5\%) splits by a deterministic hash of their identifier. No evaluated record is ever a training positive, and any future generation round inherits the same partition. The held-out evaluation set holds 32{,}442 queries over the 531 held-out indicators, about 2{,}100 per language. It is generated by a \emph{different model family} (\texttt{claude-sonnet-4-6}) under the identical protocol, so a retriever cannot score well merely by fitting the training generator's phrasing habits. The set supports overall evaluation as well as per-facet and per-language slices.

An indicator-level split does not by itself settle the leakage question, because 84.2\% of held-out records have a near-twin in training. Rather than assume the split is clean, we audit it. Section~\ref{sec:results-neardup} conditions performance on each held-out record's maximum similarity to the training split and finds performance highest on the records with no close training neighbor, so the gains are not memorization of near-twins. Cluster-level splits, in the spirit of training-data deduplication \citep{lee-etal-2022-dedup}, remain the cleaner long-term design, and we recommend them for future versions.

\subsection{LLM judge protocol}
\label{sec:judge}

Each held-out query has a single labeled positive. This ground truth is deliberately sparse, and in a near-duplicate-rich corpus it understates absolute usefulness, so the binary metrics are best read as relative comparisons between systems. To quantify the gap, an LLM judge scores retrieved (query, record) pairs on a graded 0--3 rubric. Two protocol details matter at scale. First, relevance is a property of the pair rather than of the retriever, so verdicts are cached by (query, record) and shared across all compared systems. In our English pilot this cut judge calls by 65\%. Second, within-list graded nDCG saturates when judged scores skew positive, because the ideal ranking is computed from the same skewed set. We therefore report the mean judged score of the top-$k$, which discriminates between systems. We report judged scores for the English pilot; extending the judged evaluation to all 15 languages is left to future work.

\section{Permutation-Invariant Fine-Tuning (PI-FT)}
\label{sec:method}

The method centers on a single idea---field-order permutation---supported by three complementary components: full-schema serialization, which gives permutation something to act on; similarity-guarded hard negatives; and guided contrastive training. We begin by motivating why permutation is effective, then describe the core method and each supporting component in turn.

\subsection{Why permutation augmentation induces invariance}
\label{sec:theory}

\paragraph{The goal.} We want an encoder whose output does not change when a record's fields are written in a different order. A record is a set of fields. Reordering those fields changes the text string while leaving the meaning the same, so the two strings should map to the same embedding. We can state this precisely. Let $S_n$ be the group of permutations of a record's $n$ field segments, and write $g\cdot d$ for record $d$ serialized under order $g$. The property we want is $f(g\cdot d)=f(d)$ for every permutation $g$, which is permutation invariance.

\paragraph{Why standard fine-tuning does not give this.} The obstacle is in how transformers handle order. Self-attention on its own is permutation-equivariant. Permute the input tokens and the outputs simply permute with them, so a pooled embedding does not change \citep{vaswani2017attention,lee2019settransformer}. Transformers break this symmetry on purpose, by adding positional encodings that let the model tell positions apart \citep{vaswani2017attention}. That is the right choice for ordinary text, where word order carries meaning. For a serialized record it has a side effect. The model is free to use a field's absolute position as a feature, and ordinary fine-tuning gives it no reason to avoid that. Transformers are known to take exactly this kind of shortcut, latching onto position whenever it happens to predict the target. They do so with where information sits in a long context \citep{liu2024lostmiddle}, and with the order of in-context examples \citep{lu2022fantastically}. Section~\ref{sec:results-robustness} shows the same effect for serialized records. Change the field order at indexing time and a normally fine-tuned encoder degrades sharply.

\paragraph{Why permutation augmentation removes it.} Permutation augmentation makes position uninformative. Instead of one fixed serialization, the model sees each record under a field order sampled fresh at every step, and it learns to match a query against all of these orderings. In effect it minimizes the loss averaged over orders, $\mathbb{E}_{g\sim S_n}\,\mathcal{L}\!\left(q, f(g\cdot d)\right)$. Two standard results explain why this pushes $f$ toward invariance. First, averaging any function over all orderings of its input yields a permutation-invariant function, and training on one random ordering per step is the usual stochastic estimate of that average. This is Janossy pooling, and our per-example dynamic permutation is its $\pi$-SGD form \citep{murphy2019janossy}. Second, there is a group-theoretic reading. Augmenting with all orderings of an input is equivalent to constraining the model to order-invariant functions \citep{chen2020grouptheoretic}. The averaged loss is smallest when the model scores every ordering alike. Either way, the model gains nothing from order-dependent behavior, because order no longer predicts the target.

\paragraph{What turns this into geometry.} Contrastive learning is what turns the argument into geometry. The loss pulls a query toward every serialization of its positive record, so those serializations land at nearby points and the dependence on field order fades during training. Set encoders and table encoders build this order-invariance into their architecture \citep{zaheer2017deepsets,lee2019settransformer,yang-etal-2022-tableformer}. Our encoder reaches the same property through the training data, with no change to the model. Any text embedding model can be trained this way.

\subsection{Field-order permutation and facet-protected dropout}
\label{sec:permutation}

A record $d$ with populated fields $\{(\ell_i, v_i)\}$ is serialized as labeled segments of the form ``$\ell_i$: $v_i$''. As argued above, the encoder should be invariant to the order of these segments, and standard fine-tuning does not enforce that, so we enforce it through the data. At each training step, the document is serialized under an independently sampled permutation $\pi$ of the segment order, with non-essential segments dropped at probability $p{=}0.15$:
\begin{equation}
\mathrm{ser}(d;\pi,m) \;=\; \bigoplus_{i \in \pi} m_i \cdot \text{``}\ell_i\text{: }v_i\text{''}, \qquad m_i \sim \mathrm{Bernoulli}(1-p).
\end{equation}
When the same record recurs under many orders, absolute position becomes uninformative and the field labels become the only stable cues. Minimizing the contrastive loss therefore drives the encoder toward a label-conditioned, order-invariant representation. Serialization is performed on the fly in the data loader, so every epoch presents fresh permutations of every pair at no materialization cost.

Two segments are protected from dropout. The name always survives, and so does \emph{the field a query's facet targets, within that query's positive}. Without this protection, a year-facet query would lose its time-coverage evidence 15\% of the time, which turns a training pair into a weak self-contradiction. This facet-protected dropout, where facets are defined in Section~\ref{sec:generation}, couples the augmentation to the supervision so that the pressure toward invariance never erases the signal a pair exists to teach.

The motivation for permutation is practical. Catalogs re-render metadata when portals are redesigned, federated search ingests records serialized by different producers, and downstream systems re-format before embedding. Each of these events changes the field order an index sees. Section~\ref{sec:results-robustness} measures the consequence directly under a controlled order change. A model fine-tuned \emph{without} permutation loses 7.4 nDCG@10 points, a frontier API embedding loses 6.4 zero-shot, and the permutation-trained model loses 0.2.

\subsection{Full-schema serialization}
\label{sec:serialization}

Permutation invariance requires fields to permute, and a field that is omitted from serialization cannot be retrieved against. We therefore serialize \emph{every} populated metadata field under its native schema key, without introducing a hand-curated label vocabulary. The serialized representation includes the indicator name, identifier, database, topics, unit, periodicity, definition, methodology, geographic coverage, all available time periods, and source organizations.

This design avoids the silent failure mode of hand-crafted templates. For example, omitting the methodology field makes methodology-related queries fundamentally unanswerable, yet this deficiency may remain invisible unless the evaluation explicitly contains such queries.

Serialization is also token-budget aware. Short fields are preserved in full. Geographic coverage is summarized by its positive scope---the number of economies covered together with any regional or income-group aggregates, listing individual countries only when the set is sufficiently small. The two longest fields, the definition (long form) and methodology, share the remaining token budget through max--min fair allocation. As a result, more than 99.9\% of records fit within the encoder's 512-token position-embedding limit, with a median serialized length of approximately 240 tokens. Long fields are therefore truncated deliberately according to the budget allocation rather than arbitrarily from the tail by the tokenizer.

Finally, exact country-level coverage is intentionally excluded from the serialized text. Because pooled text embeddings represent large enumerations of countries poorly, we instead treat country coverage as structured metadata to be filtered explicitly at query time.
\subsection{Hard negatives with a similarity guard}
\label{sec:negatives}

For each training query we retrieve the top 30 records with a strong \emph{multilingual} retriever (\texttt{harrier-oss-v1-0.6b}); a monolingual miner would return near-random negatives for most of the query distribution. We then skip the three highest-ranked non-positives and sample negatives from ranks 4--30. The skip window exists because the top-ranked non-positives in this corpus are disproportionately correct answers under other identifiers. A second, explicit guard drops any candidate whose document embedding has cosine similarity above 0.95 to the positive, which directly excludes the cross-database duplicates quantified in Section~\ref{sec:corpus}. Mining strength matters in both directions. Harder negatives sharpen the gradient signal \citep{xiong2021ance}, and a stronger miner makes the similarity guard's geometry more trustworthy.

\subsection{In-batch false negatives and guided training}
\label{sec:gist}

In-batch contrastive losses treat every other example in the batch as a negative \citep{henderson2017efficient}. Two properties of our regime make that assumption fail at high frequency. The first is supervision density. With 765k queries over 9{,}417 training records, each record is the positive for about 80 queries. At our training batch size $B{=}128$ the probability that a batch contains two queries sharing the same positive is about 0.58 under a birthday model (the birthday-problem approximation, treating each query's positive as uniform over the records) over the 9{,}417 training records, rising to 0.97 at $B{=}256$ and still 0.19 at $B{=}64$. For such a pair the loss pushes a query away from a document that is \emph{exactly} its positive, merely rendered under a different permutation. The second property is corpus duplication. Even across distinct records, the 84.2\% near-duplicate rate means that in-batch negatives are frequently legitimate answers.

GISTEmbed \citep{solatorio2024gist} addresses both problems with a guide model. The guide scores the batch pairs and masks from the loss any in-batch negative it considers as similar to the query as the query's own positive. Our default guide is \texttt{harrier-oss-v1-0.6b}, the strongest open multilingual embedding available to us. The guide has to judge similarity for queries in all 15 languages, and it must outrank the trainee in judgment quality, since a guide weaker than the student would filter wrongly. Cross-encoder denoising \citep{qu2021rocketqa} is the heavier alternative, run once per pair rather than once per batch. The guide-embedding formulation runs inside the training loop, and its cost scales with the guide's size: a guide comparable in size to the trainee adds modest overhead, whereas the 600M guide can increase the wall clock several-fold over an unguided run. This cost, and the observation that the strongest available guide for this corpus is a model we have already fine-tuned on it, motivate \emph{self-distillation}: using a first-generation DevData fine-tune as the guide. Section~\ref{sec:results-ablations} reports the resulting gain, which is also the cheapest guide to run.

\subsection{Training configuration}
\label{sec:training}

We fine-tune a sweep of small open bases (22--270M parameters): multilingual encoders (\texttt{multilingual-e5-small} \citep{wang2024me5}, \texttt{paraphrase-multilingual-MiniLM}, \texttt{harrier-oss-v1-270m}, \texttt{granite-embedding-97m-multilingual-r2}) and, as a controlled study of vocabulary limits, English-vocabulary encoders (NoInstruct-small in symmetric and asymmetric-pooling variants \citep{solatorio2024gist}, GIST-small, and MiniLM). Three open encoders enter only as zero-shot baselines (\texttt{jina-embeddings-v5-nano}, BGE-small \citep{xiao2023bge}, and GTE-small \citep{li2023gte}), and two proprietary API embeddings, \texttt{text-embedding-3-small} and \texttt{text-embedding-3-large}, serve as a frontier zero-shot reference. For the flagship bases we additionally fine-tune a matched control with both augmentations disabled (canonical field order, no field dropout), which supplies the no-permutation reference for the robustness test in Section~\ref{sec:results-robustness}. All models train under a cached contrastive loss with a fixed in-batch negative pool, so a base's memory footprint sets only the embedding sub-batch and never the number of negatives, which keeps the contrastive objective identical across bases. The cross-base leaderboard (Section~\ref{sec:results-main}) uses the cached-MNRL form of this loss for every base, so the comparison is on equal footing; the guided (cached-GIST) and self-distilled variants are studied separately in Section~\ref{sec:results-ablations}. Each base trains for up to 5 epochs (bf16, sequence cap 512, lr 3e-5 with 10\% warmup), with encoding conventions (e.g.\ E5 prefixes) inherited automatically from a model registry so that training and evaluation conventions can never diverge. Early stopping evaluates loss on a held-back slice of \emph{training} rows, materialized once with a fixed seed so the stopping signal is not polluted by permutation noise, and never drawn from the benchmark, with patience 5 at 1{,}500-step intervals and best-checkpoint reload. An unguided fine-tune completes in roughly half a day of single-GPU time; GIST with the 600M guide is several times slower, while a small self-distilled guide brings the guided cost back toward the unguided range. Hardware and wall-clock details are in Appendix~\ref{app:hparams}.

\section{Experiments}
\label{sec:experiments}

\subsection{Setup}
All systems index the canonical (deterministic) full-schema serialization, using exact inner-product search over L2-normalized embeddings, so that differences are attributable to the embeddings alone. The baselines are the zero-shot open models listed above and OpenAI \texttt{text-embedding-3-small} and \texttt{text-embedding-3-large} as frontier API references. We report R@$k$, MRR, and nDCG@10 against the labeled positive, both overall and per language, with significance assessed by a paired bootstrap over 10{,}000 resamples. One naming convention runs throughout: on the cross-base leaderboard ``e5-ft'' is the unguided cached-MNRL model (Table~\ref{tab:main}), while the single-best ``self-distilled flagship'' (cached-GIST, Section~\ref{sec:results-ablations}) is used only in the unseen-language results (Section~\ref{sec:results-unseen}). The English-only benchmark from our pilot, 303 queries authored against a different serialization, serves as an out-of-distribution check.

\subsection{Main results}
\label{sec:results-main}

Table~\ref{tab:main} reports the multilingual holdout. The fine-tuned 118M \texttt{multilingual-e5-small} leads every system, beating the frontier API embedding \texttt{text-embedding-3-large} by 15.1 nDCG@10 points (0.707 vs 0.556) and 15.6 R@10 points, at a fraction of the size and with no per-query cost. \texttt{harrier-270m-ft} matches it (0.706), and a third multilingual base, \texttt{paraphrase-ml-minilm-ft}, also clears the API flagship (0.678). The 33M English-vocabulary base trails the multilingual ones (0.483) but still beats \texttt{text-embedding-3-small}; Section~\ref{sec:results-tokenizer} attributes its ceiling to vocabulary. The top three are statistically tied: a paired bootstrap over the 32{,}442 queries places \texttt{multilingual-e5-small-ft} only 0.002 nDCG@10 above \texttt{harrier-270m-ft} (95\% CI $[-0.001, +0.004]$, $p{=}0.13$). The strongest zero-shot baseline is open, \texttt{jina-embeddings-v5-nano} at 0.561, marginally above \texttt{text-embedding-3-large}; it, the API embedding, and the 600M \texttt{harrier-oss-v1-0.6b} (0.530) all fall about 0.15 nDCG@10 short of the fine-tuned models. Fine-tuning roughly doubles most bases over their zero-shot scores (e5 $0.367\rightarrow0.707$, paraphrase $0.304\rightarrow0.678$), which is the evidence that supervision dominates parameter count here. A BM25 lexical baseline reaches 0.413 nDCG@10, above the zero-shot e5 base (0.367) but well below every fine-tuned model; it is more competitive on the English pilot (0.478) and near-perfect on identifier lookup (Section~\ref{sec:results-facets}), which is why a lexical channel remains a useful complement for exact codes rather than a replacement for the encoder.

\begin{table}[t]
\caption{Multilingual holdout retrieval: 32{,}442 queries over 531 held-out indicators, raw-schema serialization. Fine-tuned (ft) models use permutation-invariant fine-tuning under the cached-MNRL loss, on equal footing across bases; the guided and self-distilled variants are in Section~\ref{sec:results-ablations} (Table~\ref{tab:guide}). The API embeddings are zero-shot.}
\label{tab:main}
\begin{center}\small
\begin{tabular}{llcccc}
\toprule
Model & type & R@1 & R@10 & MRR & nDCG@10 \\
\midrule
multilingual-e5-small-ft (118M) & ft, multilingual & 0.522 & \textbf{0.891} & 0.652 & \textbf{0.707} \\
harrier-270m-ft (270M) & ft, multilingual & \textbf{0.528} & 0.881 & \textbf{0.653} & 0.706 \\
paraphrase-ml-minilm-ft (118M) & ft, multilingual & 0.476 & 0.879 & 0.616 & 0.678 \\
granite-97m-ft (97M) & ft, multilingual & 0.374 & 0.747 & 0.501 & 0.556 \\
noinstruct-small-ft (33M) & ft, English-vocab & 0.304 & 0.680 & 0.427 & 0.483 \\
\midrule
jina-embeddings-v5-nano & zero-shot open & 0.397 & 0.731 & 0.511 & 0.561 \\
harrier-oss-v1-0.6b (600M) & zero-shot open & 0.369 & 0.701 & 0.481 & 0.530 \\
text-embedding-3-large & zero-shot API & 0.386 & 0.735 & 0.498 & 0.556 \\
text-embedding-3-small & zero-shot API & 0.253 & 0.520 & 0.346 & 0.381 \\
BM25 (lexical) & lexical & 0.360 & 0.474 & 0.402 & 0.413 \\
\bottomrule
\end{tabular}
\end{center}
\end{table}

Per-language results (Table~\ref{tab:perlang}) show that the fine-tuned multilingual encoders are near-uniform across all seven scripts. \texttt{multilingual-e5-small-ft} ranges only from 0.827 (Swahili) to 0.921 (English), an 0.09 spread, with English barely ahead of Chinese (0.914). This is the evidence that the English-heavy training mix (English is the document language) does not compromise multilingual parity: if it had, the low-resource tail would lag far behind. Against the frontier API embedding \texttt{text-embedding-3-large} (last row), the fine-tune's edge is smallest on English ($+0.03$ R@10) and largest on the low-resource languages (Hindi $+0.28$, Swahili $+0.19$, Arabic $+0.12$), where the API itself is weak, and the full 15-language results show the same pattern. Fine-tuning therefore helps most exactly the languages a general embedding serves worst. The English-vocabulary base behaves very differently, which Section~\ref{sec:results-tokenizer} takes up.

\begin{table}[t]
\caption{Per-language R@10 on the multilingual holdout, representative languages across script families. Multilingual bases are near-uniform; the English-vocabulary base collapses on non-Latin scripts. Per-language results for all 15 languages accompany the released evaluation outputs.}
\label{tab:perlang}
\begin{center}\small
\begin{tabular}{lcccccccc}
\toprule
Model & en & de & ru & ar & hi & sw & zh & ja \\
\midrule
multilingual-e5-small-ft & \textbf{0.921} & \textbf{0.873} & \textbf{0.891} & \textbf{0.878} & 0.895 & \textbf{0.827} & \textbf{0.914} & \textbf{0.880} \\
harrier-270m-ft & 0.916 & 0.863 & 0.884 & 0.863 & \textbf{0.899} & 0.799 & 0.908 & 0.861 \\
granite-97m-ft & 0.888 & 0.734 & 0.776 & 0.711 & 0.778 & 0.493 & 0.808 & 0.772 \\
noinstruct-small-ft (English-vocab) & 0.912 & 0.723 & 0.654 & 0.597 & 0.603 & 0.637 & 0.395 & 0.450 \\
\midrule
text-embedding-3-large (zero-shot) & 0.889 & 0.805 & 0.797 & 0.758 & 0.613 & 0.639 & 0.828 & 0.784 \\
\bottomrule
\end{tabular}
\end{center}
\end{table}

\subsection{Where the base vocabulary limits multilingual transfer}
\label{sec:results-tokenizer}

Before training, we audited what each base's tokenizer preserves of the multilingual queries (Table~\ref{tab:tokenizer}). For an English WordPiece vocabulary \citep{devlin2019bert}, Latin-script languages tokenize cleanly (0\% UNK; rich subword diversity); Arabic-script, Indic, and Cyrillic queries survive as \emph{character-level} sequences (0.5--3\% UNK but only $\sim$130--170 distinct tokens, 26--49 tokens per query); Japanese loses 34\% and Chinese 68\% of tokens to UNK outright. Embedding-space distinctness degrades in the same order. This yields a falsifiable prediction, fine-tuning gains for English-vocabulary bases should be near-full on Latin scripts, partial on character-level scripts, and minimal on Chinese, which the per-language results test directly. The prediction holds (Table~\ref{tab:perlang}). After identical fine-tuning, the English-vocabulary NoInstruct base reaches 0.912 R@10 on English but only 0.395 on Chinese and 0.450 on Japanese, with Latin-script German in between at 0.723, while the multilingual bases stay above 0.86 on Chinese and Japanese alike. The collapse tracks the UNK and token-distinctness tiers of Table~\ref{tab:tokenizer}. The audit generalizes a point made for generation costs \citep{petrov2023tokenizers} to retrieval capability: the tokenizer sets a per-language information ceiling that fine-tuning cannot lift.

\begin{table}[t]
\caption{Tokenizer audit of an English WordPiece vocabulary on the multilingual training queries (150 queries per language). UNK\%: fraction of tokens mapped to the unknown token. Distinct: unique token types observed. Mean pairwise cosine of base-encoder embeddings indicates how distinguishable same-language queries remain (English baseline 0.63).}
\label{tab:tokenizer}
\begin{center}
\small
\begin{tabular}{lcccc}
\toprule
Script tier & Languages & UNK\% & Distinct tokens & Pairwise cos \\
\midrule
Latin & es, fr, de, pt, id, sw, tr & 0.0 & 529--697 & 0.73--0.77 \\
Character-level & ru, ar, hi, bn, ur & 0.0--2.8 & 126--170 & 0.81--0.87 \\
Lossy & ja & 34.3 & 259 & 0.81 \\
 & zh & 68.3 & 121 & 0.76 \\
\bottomrule
\end{tabular}
\end{center}
\end{table}

\subsection{Facet analysis}
\label{sec:results-facets}

Different facets exercise different parts of the schema, so per-facet performance tests whether full-schema serialization actually makes field-targeted queries work. Table~\ref{tab:facet} reports per-facet R@10 for the two leading fine-tunes. The field-targeted facets that motivate full-schema serialization are all well served: geographic (0.919), year (0.924), combined geo-year (0.951), methodology (0.914), unit (0.856), and frequency (0.885) for \texttt{multilingual-e5-small-ft}. Identifier lookup (code) is near-perfect at 0.989, and BM25 is marginally higher still on that facet (0.997), so a lexical fallback remains the production-safe complement for exact codes. The two weakest facets are database (0.702) and thematic (0.735); both are inherently low-precision, since many indicators share a database and a thematic query names a broad topic rather than a specific series. The ordering is stable across models. Measured against the zero-shot e5 base, fine-tuning lifts the field-targeted facets most: geo-year by 0.50, geographic by 0.44, year by 0.42, and frequency by 0.37, against only 0.26--0.27 for the semantic facets (definition, methodology) that the base already handles. So the framework adds the most exactly where a query targets a specific field, and least where plain concept matching already suffices. The advantage over the frontier API embedding \texttt{text-embedding-3-large} (last row) is concentrated in the field-targeted facets, geo-year ($+0.21$), frequency ($+0.20$), source ($+0.18$), geographic ($+0.16$), and year ($+0.15$), and is smallest on the semantic facets, definition ($+0.07$) and keyword and natural ($+0.11$). Identifier lookup is the extreme case ($+0.45$), since exact codes are not a semantic-similarity task.

\begin{table}[t]
\caption{Per-facet R@10 on the multilingual holdout, the two leading fine-tuned models. Field-targeted facets (geo, year, methodology, unit, frequency) are well served; database and thematic are the hardest. Gains over the zero-shot base concentrate on the field-targeted facets (see text).}
\label{tab:facet}
\begin{center}\footnotesize
\setlength{\tabcolsep}{4pt}
\begin{tabular}{lccccccccccccc}
\toprule
Model & code & def & geo & geo+yr & year & meth & unit & freq & kw & nat & src & db & them \\
\midrule
e5-small-ft & \textbf{0.989} & \textbf{0.964} & \textbf{0.919} & \textbf{0.951} & \textbf{0.924} & \textbf{0.914} & \textbf{0.856} & \textbf{0.885} & \textbf{0.965} & \textbf{0.966} & \textbf{0.860} & 0.702 & \textbf{0.735} \\
harrier-270m-ft & 0.953 & 0.959 & 0.911 & 0.933 & 0.918 & 0.907 & 0.849 & 0.870 & 0.950 & 0.955 & 0.852 & \textbf{0.706} & 0.725 \\
\midrule
text-embedding-3-large (0-shot) & 0.544 & 0.891 & 0.762 & 0.745 & 0.770 & 0.787 & 0.719 & 0.687 & 0.853 & 0.855 & 0.679 & 0.575 & 0.581 \\
\bottomrule
\end{tabular}
\end{center}
\end{table}

\begin{figure}[t]
\centering
\includegraphics[width=0.92\linewidth]{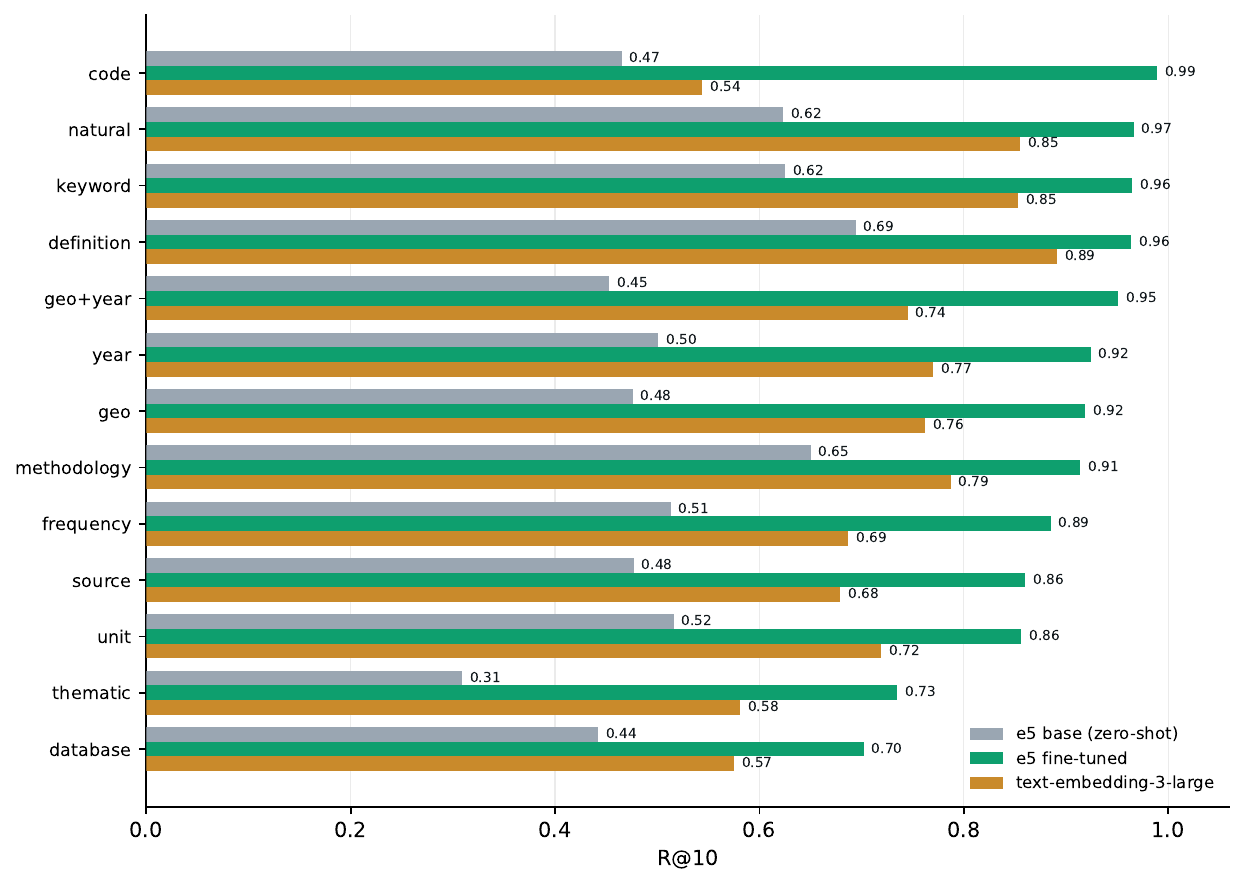}
\caption{Per-facet R@10 on the multilingual holdout for the base encoder, the fine-tuned e5 (cached-MNRL, matching Table~\ref{tab:facet}), and the frontier API embedding, sorted by the fine-tune. Fine-tuning lifts the field-targeted facets most; database and thematic remain the hardest for every system.}
\label{fig:facet}
\end{figure}

\subsection{Robustness to serialization order}
\label{sec:results-robustness}

We rebuild the index under a different field order and re-evaluate; nothing else changes, so any drop is pure order sensitivity. Table~\ref{tab:robust} shows the result. Every permutation-trained model is order-invariant to within noise: the flagship \texttt{multilingual-e5-small-ft} loses 0.2 nDCG@10 points, \texttt{harrier-270m-ft} essentially zero, and the largest drop among the fine-tuned models, \texttt{granite-97m-ft}, is 1.1 points. The zero-shot API embeddings, by contrast, lose 6.4 (\texttt{text-embedding-3-large}) and 8.5 (\texttt{text-embedding-3-small}) points to the same change. The matched no-permutation controls isolate the cause. Trained on the same base and data but with a fixed field order, the fine-tunes lose 7.4 (e5), 6.5 (harrier-270m), and 14.2 (granite) nDCG@10 points under the order change, against 0.2, 0.0, and 1.1 for their permutation-trained counterparts. Permutation training is also no worse on the canonical index. It is 1.6 to 2.8 points better across the three bases. So the invariance comes from the permutation augmentation rather than from fine-tuning in general, and it costs nothing in-distribution. Figure~\ref{fig:robustfull} (appendix) shows the same canonical-versus-permuted comparison across all 22 systems, including every zero-shot baseline: the permutation-trained fine-tunes collapse to a point while every other model, open or API, slides left.

\begin{table}[t]
\caption{Robustness to an index field-order change: nDCG@10 on the canonical vs a permuted-order index (holdout). $\Delta$ is in nDCG@10 points. Permutation-trained models are invariant to within noise; the no-permutation controls and the zero-shot API embeddings are not. No-permutation controls are shown for all three multilingual bases.}
\label{tab:robust}
\begin{center}\small
\begin{tabular}{lccc}
\toprule
Model & canonical & permuted & $\Delta$ \\
\midrule
\multicolumn{4}{l}{\emph{Permutation-trained fine-tunes}} \\
harrier-270m-ft & 0.706 & \textbf{0.706} & $-0.0$ \\
multilingual-e5-small-ft & \textbf{0.707} & 0.705 & $-0.2$ \\
paraphrase-ml-minilm-ft & 0.678 & 0.679 & $+0.2$ \\
noinstruct-small-ft & 0.483 & 0.481 & $-0.2$ \\
minilm-l6-ft & 0.453 & 0.451 & $-0.2$ \\
granite-97m-ft & 0.556 & 0.545 & $-1.1$ \\
\midrule
\multicolumn{4}{l}{\emph{No-permutation control (same base and data)}} \\
multilingual-e5-small, no permutation & 0.679 & 0.605 & $-7.4$ \\
harrier-270m, no permutation & 0.690 & 0.625 & $-6.5$ \\
granite-97m, no permutation & 0.540 & 0.398 & $-14.2$ \\
\midrule
\multicolumn{4}{l}{\emph{Zero-shot API}} \\
text-embedding-3-large & 0.556 & 0.492 & $-6.4$ \\
text-embedding-3-small & 0.381 & 0.297 & $-8.5$ \\
\bottomrule
\end{tabular}
\end{center}
\end{table}

\begin{figure}[t]
\centering
\includegraphics[width=0.78\linewidth]{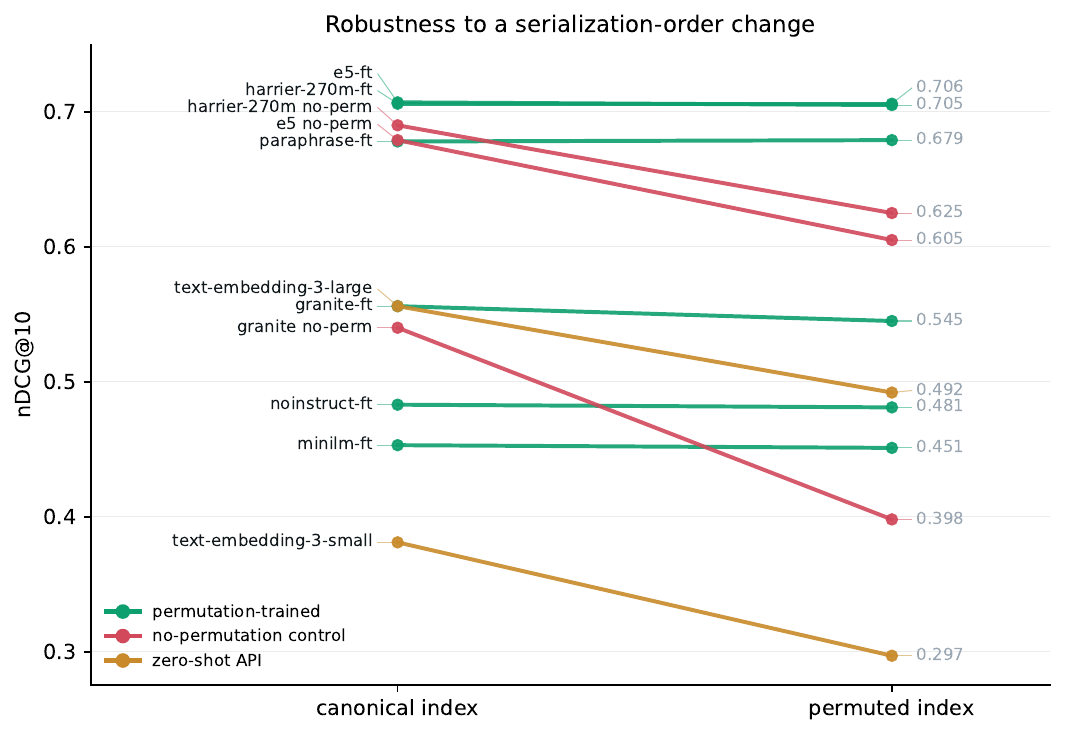}
\caption{nDCG@10 on the canonical versus a permuted-order index. Permutation-trained fine-tunes stay flat; the no-permutation controls and the zero-shot API embeddings fall.}
\label{fig:robustness}
\end{figure}

\subsection{Near-duplicate conditioning}
\label{sec:results-neardup}

The 84.2\% near-duplicate rate (Section~\ref{sec:splits}) raises the worry that the gains are memorization of training twins, and the conditioning here rules that out. We split the held-out queries by whether their target indicator has a training-set neighbor at base-encoder cosine $\geq 0.95$ (computed with NoInstruct-small, the same flag behind the 84.2\%) and compare R@10 within each bucket (Table~\ref{tab:neardup}). Every multilingual fine-tune scores higher on the \emph{distinct} records, those with no close training neighbor, than on the near-duplicates: \texttt{multilingual-e5-small-ft} reaches 0.911 against 0.886 and \texttt{harrier-270m-ft} 0.933 against 0.871. Memorization of twins would produce the opposite ordering. The zero-shot base shows the largest gap (e5 $+0.22$), so the duplicate records are harder for every system and fine-tuning does not lean on them. The result is robust to how a duplicate is defined, across two embedding encoders and a lexical criterion (Appendix~\ref{app:neardup}). The English-vocabulary fine-tunes are the exception, with a small negative gap (about $-0.02$ to $-0.04$), so the claim holds specifically for the multilingual models.

The same conclusion holds under a stricter, dedup-aware split. Clustering the 9{,}948 indicators by near-duplication (connected components at cosine $\geq 0.95$) yields 2{,}688 clusters, one as large as 1{,}174 indicators, and 84.2\% of held-out indicators share a cluster with a training indicator, matching the pairwise rate. The 84 held-out indicators (5{,}148 queries) whose cluster contains no training member, the records a dedup-aware split would isolate, are if anything easier: the flagship reaches 0.93 R@10 on them against 0.90 on the records that share a cluster with training, and the base e5 0.70 against 0.48. Retrieval therefore does not depend on near-duplicate leakage. A full retrain on a clustered split is the gold-standard version of this control and remains future work. The concern in any case bears only on absolute retrieval quality: the order-invariance result (Section~\ref{sec:results-robustness}) is a within-evaluation comparison of the same queries and records under two field orders, so any leakage affects both conditions equally and cannot create or mask an order-sensitivity gap.

\begin{table}[t]
\caption{Near-duplicate conditioning on the multilingual holdout. Held-out queries are split by whether their target indicator has a training neighbor at NoInstruct-small cosine $\geq 0.95$ (84.2\% of indicators are near-duplicates: 27{,}170 queries near-dup, 5{,}272 distinct). Every multilingual fine-tune scores higher on the distinct records.}
\label{tab:neardup}
\begin{center}\small
\begin{tabular}{lccc}
\toprule
Model & R@10 near-dup & R@10 distinct & $\Delta$ \\
\midrule
multilingual-e5-small-ft & \textbf{0.886} & 0.911 & $+0.025$ \\
harrier-270m-ft & 0.871 & \textbf{0.933} & $+0.062$ \\
granite-97m-ft & 0.735 & 0.809 & $+0.074$ \\
\midrule
multilingual-e5-small (0-shot) & 0.482 & 0.698 & $+0.216$ \\
\bottomrule
\end{tabular}
\end{center}
\end{table}

\subsection{Generalization to unseen languages}
\label{sec:results-unseen}

The 15 supervised languages raise the question of whether fine-tuning helps or hurts on languages it never saw. We test this directly. Using the same grounded protocol and the \texttt{claude-sonnet-4-6} generator, we produce 240 queries per language for held-out indicators in ten languages absent from training: five across South and Southeast Asia and adjacent regions (Tagalog, Vietnamese, Korean, Thai, Persian) and five lower-resource African languages (Kinyarwanda, Yoruba, Hausa, Amharic, Zulu), together adding three scripts the training set never contained (Hangul, Thai, and Ethiopic); Persian is a new language in the already-covered Arabic script. Table~\ref{tab:unseen} reports R@10 for the base encoder, its fine-tune, and the frontier API embedding.

Fine-tuning improves over the base on every one of the ten languages, by 0.32 on average for the first group and 0.20 for the African group, so the framework transfers rather than forgets, even to unseen scripts (Amharic gains 0.40). Here e5-ft is the self-distilled flagship (Section~\ref{sec:results-ablations}); the cached-MNRL leaderboard model transfers similarly (means 0.83 and 0.50). The fine-tuned 118M model also beats \texttt{text-embedding-3-large} on these unseen languages, and the margin is largest where it matters most. On the African set the API embedding falls below even the base \texttt{multilingual-e5-small} (0.31 vs 0.34 mean) and reaches only 0.21 on Ethiopic-script Amharic, while the fine-tune reaches 0.80. Absolute scores on the hardest African languages stay modest and depend on the generator's own quality in those languages, but the queries are identical across models, so the comparison is fair. This extends the equity finding of Section~\ref{sec:results-main} beyond the supervised set: the framework helps most for the languages a general-purpose embedding serves worst, whether or not they appear in training.

\begin{table}[t]
\caption{Generalization to languages absent from fine-tuning: R@10 over 240 grounded queries per language for held-out indicators. e5-ft is the self-distilled flagship (cached-GIST guided by a first-generation fine-tune, Section~\ref{sec:results-ablations}); the API embedding is zero-shot. Fine-tuning improves over the base on all ten languages and exceeds the API embedding, most on the lower-resource African set, where the API falls below the base and collapses on Ethiopic-script Amharic.}
\label{tab:unseen}
\begin{center}\small
\begin{tabular}{lcccc}
\toprule
Language (script) & base & e5-ft & 3-large & ft$-$base \\
\midrule
\multicolumn{5}{l}{\emph{South/Southeast Asia and adjacent}} \\
Tagalog (Latin) & 0.717 & \textbf{0.908} & 0.887 & $+0.19$ \\
Vietnamese (Latin) & 0.450 & \textbf{0.829} & 0.721 & $+0.38$ \\
Korean (Hangul) & 0.412 & \textbf{0.738} & 0.700 & $+0.33$ \\
Thai (Thai) & 0.575 & \textbf{0.900} & 0.650 & $+0.33$ \\
Persian (Arabic) & 0.446 & \textbf{0.833} & 0.688 & $+0.39$ \\
\emph{mean} & 0.520 & \textbf{0.842} & 0.729 & $+0.32$ \\
\midrule
\multicolumn{5}{l}{\emph{Lower-resource African}} \\
Amharic (Ethiopic) & 0.396 & \textbf{0.796} & 0.208 & $+0.40$ \\
Zulu (Latin) & 0.407 & \textbf{0.542} & 0.394 & $+0.14$ \\
Hausa (Latin) & 0.292 & \textbf{0.471} & 0.271 & $+0.18$ \\
Kinyarwanda (Latin) & 0.339 & \textbf{0.504} & 0.369 & $+0.17$ \\
Yoruba (Latin) & 0.271 & \textbf{0.408} & 0.312 & $+0.14$ \\
\emph{mean} & 0.341 & \textbf{0.544} & 0.311 & $+0.20$ \\
\bottomrule
\end{tabular}
\end{center}
\end{table}

\begin{figure}[t]
\centering
\includegraphics[width=0.92\linewidth]{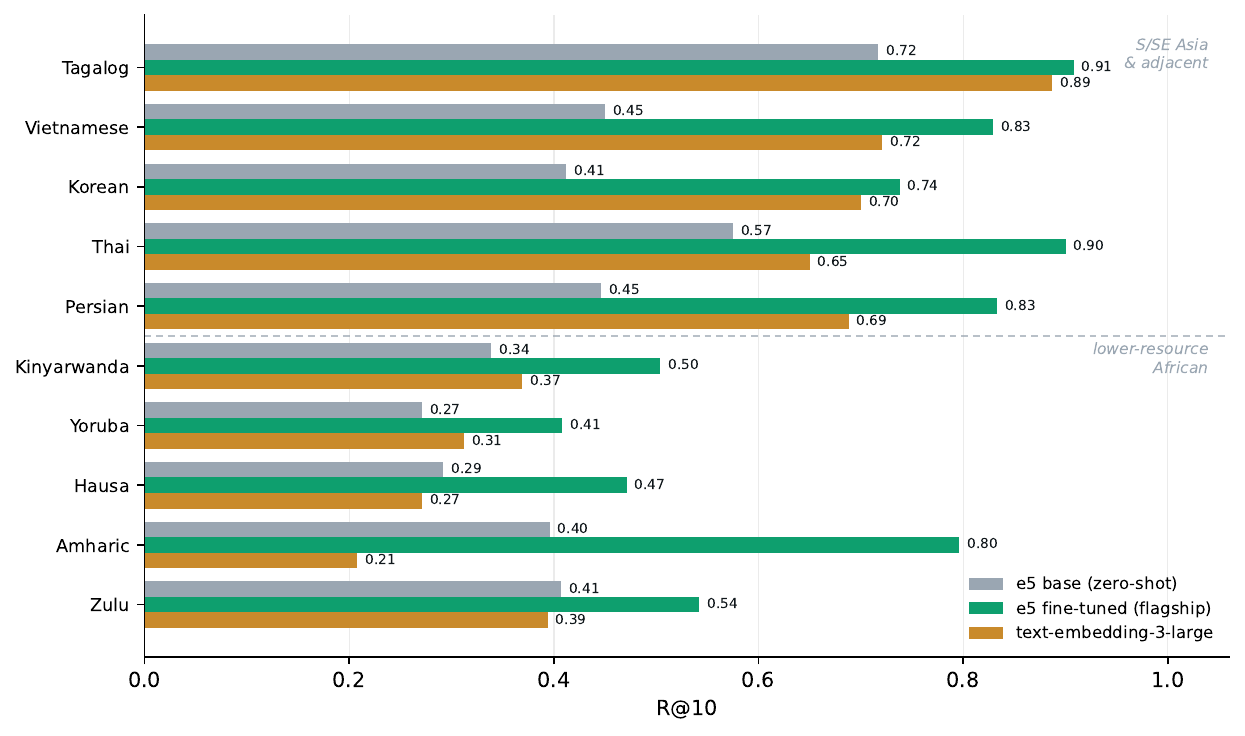}
\caption{Generalization to languages absent from fine-tuning: R@10 for the base encoder, the self-distilled flagship fine-tune, and the frontier API embedding, across five South and Southeast Asian and adjacent languages and five lower-resource African languages. The fine-tune leads on all ten; the API embedding falls below the base on the African set and to 0.21 on Amharic.}
\label{fig:unseen}
\end{figure}

\subsection{Ablations}
\label{sec:results-ablations}

\paragraph{Pooling (NoInstruct).} Symmetric mean pooling reaches 0.483 nDCG@10, above the model card's prescribed asymmetric pooling (mean queries, CLS documents) at 0.471. So the card's pooling does not help once the encoder is fine-tuned, and the result reproduces the pilot (symmetric ahead by a similar margin).

\paragraph{Guide model and self-distillation.} The cross-base leaderboard (Table~\ref{tab:main}) trains every model with the unguided cached-MNRL loss. Adding a GIST guide, which masks in-batch false negatives, improves \texttt{multilingual-e5-small} further. All rows of Table~\ref{tab:guide} share the same configuration (five epochs, sequence cap 512) and differ only in the guide, so the comparison is controlled. An independent open guide (\texttt{harrier-oss-v1-0.6b}) lifts nDCG@10 from 0.707 to 0.722. Replacing it with a \emph{self-distilled} guide, a first-generation DevData fine-tune of the same base, helps only when that seed is strong enough: a guide trained with cached-MNRL gives 0.714, below the independent guide, while a guide trained with cached-GIST gives 0.730, above it. Guide quality therefore sets student quality, and self-distillation can be iterated once a generation clears the independent-guide bar. The gaps are small but not noise: a paired bootstrap over the 32{,}442 queries makes every adjacent ordering in Table~\ref{tab:guide} significant ($p<0.001$), including the independent guide over the cached-MNRL self-distill and the cached-GIST self-distill over the independent guide. Order-invariance survives guiding throughout, with order-change penalties between 0.2 and 1.4 nDCG@10 points, so the gain does not cost the property the framework exists to provide. A practical bonus is cost: the self-distilled guide is a 118M model rather than the 600M default, which removes most of GIST's wall-clock overhead.

\begin{table}[t]
\caption{GIST guide for \texttt{multilingual-e5-small} (multilingual holdout, raw-schema serialization). All rows share one configuration (five epochs, sequence cap 512) and differ only in the guide. Self-distillation uses a first-generation DevData fine-tune as the guide. $\Delta$ is the order-change penalty of Section~\ref{sec:results-robustness}. The unguided row is the Table~\ref{tab:main} flagship.}
\label{tab:guide}
\begin{center}\small
\begin{tabular}{lcc}
\toprule
Guide & nDCG@10 & $\Delta$ order \\
\midrule
none (unguided cached-MNRL flagship) & 0.707 & $-0.2$ \\
independent open guide (\texttt{harrier-0.6b}) & 0.722 & $-1.4$ \\
self-distilled, first-gen cached-MNRL guide & 0.714 & $-0.2$ \\
self-distilled, first-gen cached-GIST guide & \textbf{0.730} & $-0.6$ \\
\bottomrule
\end{tabular}
\end{center}
\end{table}

Facet-protected dropout, miner strength, and the similarity guard are not separately ablated here and are left to future work; the permutation ablation itself is the no-permutation control of Section~\ref{sec:results-robustness}.

\section{Discussion}
\label{sec:discussion}

\paragraph{Findability is where development-data value is lost.}
The cost of producing most development statistics has already been paid, and the loss occurs at discovery, when an analyst cannot find a series that exists and so re-collects it or proceeds without evidence \citep{worldbank2021wdr,chapman2020dataset}. Findability is the first FAIR principle \citep{wilkinson2016fair}, yet in practice it has meant English keyword search over titles. A multilingual, schema-aware retriever widens it: the same catalog becomes searchable by a francophone ministry analyst, a Swahili-speaking journalist, or a Bengali student in their own words, and against the methodology, coverage, and unit fields that titles never expose. Our per-language results show this widening is also an equity gain. The fine-tune's margin over the frontier API embedding is largest on the low-resource, non-Latin languages those users speak (Table~\ref{tab:perlang}), so the benefit accrues most to the people a general-purpose embedding serves worst.

\paragraph{A small self-hosted encoder fits the institutional setting.}
LLM applications retrieve and then read \citep{lewis2020rag}, and agents expose catalogs as tools, so the embedding decides whether an answer is grounded in the right statistic, and it is invoked on every query, which is where cost and data governance bind hardest. Statistical agencies often cannot route metadata or query streams through external APIs, per-query pricing sits poorly with public-sector budgets, and serving energy scales with model size \citep{strubell2019energy,schwartz2020green}, a burden heaviest for the lower-income institutions development data most needs to serve. Our results meet these constraints empirically: with domain-targeted supervision a 118M encoder runs on a CPU yet outperforms zero-shot systems many times its size, so for specialized corpora supervision quality dominates parameter count. Order-invariance is what keeps such a self-hosted model dependable, because these catalogs re-render metadata after portal redesigns and ingest records serialized by different producers, and a model that reads the field labels keeps working across those changes.

\paragraph{LLM-generated supervision and its audits.}
Every label in this work is LLM-generated, at a cost three orders of magnitude below human annotation at the same scale. Generation also buys coverage a log cannot give: queries reach every field in every language, including phrasings a real catalog has never received, so the encoder is trained for the questions users might ask rather than only the ones already asked. We do not claim that synthetic data is automatically trustworthy. Its value here is that its failure modes are specific and measurable. Those failure modes are generator-style leakage, split leakage through near-duplicates, ungrounded constraints, and coverage gaps. Each one gets an explicit control or audit, namely generator separation, similarity audits, grounding rules, and per-language and per-facet coverage accounting. The audits themselves produced findings, including the near-duplicate rate and the tokenizer tiers, that went on to shape the method.

\paragraph{Self-distillation compounds domain supervision.}
For masking false negatives in this corpus, a model already fine-tuned on it makes a stronger guide than any general-purpose embedding we tried. Using a first-generation DevData encoder as the GIST guide improves the next generation (Section~\ref{sec:results-ablations}), and the better the guide, the better the student, so the procedure can in principle be iterated. Two caveats keep this honest. The gain over the unguided flagship is real but modest, and it makes the framework a bootstrap: the best configuration is no longer reproducible from public base models in a single pass, since it depends on a prior fine-tune. We therefore report the unguided cross-base leaderboard as the primary, self-contained result and self-distillation as an enhancement for practitioners who can afford a second training round. The bonus is operational rather than only statistical: a 118M self-distilled guide removes most of the wall-clock cost that a 600M guide imposes.

\paragraph{Composition with structured search.}
Hard constraints deserve enforcement rather than embedding. A production deployment should still parse confident year and geography mentions into filters over the same structured fields our serializer reads \citep{tunkelang2009faceted,guo2009nerq}, fuse lexical evidence for codes and acronyms \citep{robertson2004bm25f,cormack2009rrf}, and let the schema-invariant encoder carry the semantic remainder. What the encoder changes is the floor. Structured queries keep working when the parser misses, in any of the 15 languages.

\section{Limitations}
\label{sec:limitations}


All queries are LLM-generated. Using two independent generators reduces style leakage between training and evaluation, but it does not address whether human search behavior differs systematically from generated queries. Validation against real portal logs therefore remains an important direction for future work. Human evaluation of query quality is also still needed for lower-resource languages, where the generator performs worst and automatic grounding is difficult because translated geographic names often defeat string-matching methods.

Ground truth consists of a single labeled positive per query in a corpus containing many near-duplicates, so binary retrieval metrics should be interpreted as relative rather than absolute. In addition, the judged evaluation protocol has so far been applied only to the English pilot.

All experiments were conducted with a single random seed on a single catalog. Although the methodology should generalize to any dataset with labeled metadata fields, we have not yet demonstrated this empirically. Evaluating a second catalog from a different domain is our highest-priority test of external validity and the main planned next step.

The supervised training data cover seven writing systems but still omit most of the world's languages. As shown in Section~\ref{sec:results-unseen}, the training framework transfers successfully to ten unseen languages spanning three additional scripts rather than catastrophically forgetting them. However, performance on the lowest-resource African languages remains modest and is ultimately limited by the quality of the query generator itself.

Our metadata serialization also omits indicator dimensions (e.g., disaggregation dimensions and their values). This design keeps the input representation simple and broadly applicable across catalogs, but it also prevents the model from exploiting potentially informative structured metadata. Incorporating serialized dimensions while preserving permutation invariance is a natural direction for future work.

Self-distillation is inherently a bootstrapping procedure rather than a single-pass training framework because its best-performing configuration relies on an initial fine-tuned model to generate improved supervision. Finally, the cross-base comparison fixes the training objective to cached-MNRL, meaning the leaderboard compares backbone models under a common objective rather than under each model's individually optimized objective.

\section{Conclusion}

Serializing a structured record imposes an arbitrary field order, and once an encoder is fine-tuned, that hidden design choice can substantially affect retrieval quality. We show that this dependence can be removed with a simple permutation-invariant fine-tuning framework requiring only a two-line modification to the data loader, reducing the order-change penalty from 7.4 nDCG@10 points to nearly zero without sacrificing in-distribution performance.

We demonstrate the approach on multilingual retrieval of development statistics. Combined with facet-targeted query generation grounded in record content, duplication-aware hard negatives, and guided contrastive training, a 118M-parameter open embedding model achieves retrieval quality comparable to or better than systems orders of magnitude larger while remaining robust to arbitrary serialization order. Guided training further improves performance when the guide is a first-generation model trained with the same framework, raising the best model to 0.730 nDCG@10 while preserving permutation invariance.

Finally, the complete pipeline—from synthetic data generation through training, evaluation, and robustness audits—is made open. By releasing the full workflow, we aim to make high-quality multilingual retrieval practical for any catalog built from structured metadata rather than only for organizations with access to large proprietary models or expensive training infrastructure.

\subsubsection*{Ethics statement}
The corpus is public institutional metadata, and no personal data is processed. Synthetic supervision inherits the generator's biases in topic emphasis and phrasing across languages, and generation quality is plausibly lower for lower-resource languages, so deployments serving such users should add human review before relying on per-language parity claims. Improved discoverability of statistics is broadly beneficial but not neutral. Surfacing an indicator without its methodological caveats can enable misuse, which is why our serialization deliberately retains the methodology and source fields, so that retrieval surfaces context alongside numbers.

\subsubsection*{Reproducibility statement}
All code (generation, mining, training, evaluation, and audits), the benchmark, the prompts (Appendix~\ref{app:prompts}), the hyperparameters (Appendix~\ref{app:hparams}), and the fine-tuned checkpoints are released. Every dataset artifact is regenerable from the public catalog at low API cost. The indicator split is a deterministic hash, and all experiments run from two commands documented in the repository.

\subsubsection*{Acknowledgments}
This work is supported by the “World Bank Group President's Innovation Awards - FY26” project funded by The World Bank - P514969.

\subsubsection*{Disclaimer and disclosure of AI use}
The findings, interpretations, and conclusions expressed in this paper are entirely those of the authors. They do not necessarily represent the views of the International Bank for Reconstruction and Development/World Bank and its affiliated organizations, or those of the Executive Directors of the World Bank or the governments they represent.

This work used AI tools at various stages, including open-source AI models for embedding inference and fine-tuning. In addition, Claude was employed to enhance the manuscript’s readability and to accelerate the development of the experiment pipeline.

\bibliography{references}

@inproceedings{vaswani2017attention,
  title     = {Attention Is All You Need},
  author    = {Vaswani, Ashish and Shazeer, Noam and Parmar, Niki and Uszkoreit, Jakob and Jones, Llion and Gomez, Aidan N. and Kaiser, {\L}ukasz and Polosukhin, Illia},
  booktitle = {Advances in Neural Information Processing Systems (NeurIPS)},
  year      = {2017}
}

@inproceedings{reimers-gurevych-2019-sentence,
  title     = {Sentence-{BERT}: Sentence Embeddings using {S}iamese {BERT}-Networks},
  author    = {Reimers, Nils and Gurevych, Iryna},
  booktitle = {Proceedings of the 2019 Conference on Empirical Methods in Natural Language Processing (EMNLP-IJCNLP)},
  year      = {2019}
}

@inproceedings{gao-etal-2021-simcse,
  title     = {{S}im{CSE}: Simple Contrastive Learning of Sentence Embeddings},
  author    = {Gao, Tianyu and Yao, Xingcheng and Chen, Danqi},
  booktitle = {Proceedings of the 2021 Conference on Empirical Methods in Natural Language Processing (EMNLP)},
  year      = {2021}
}

@inproceedings{thakur2021beir,
  title     = {{BEIR}: A Heterogeneous Benchmark for Zero-shot Evaluation of Information Retrieval Models},
  author    = {Thakur, Nandan and Reimers, Nils and R{\"u}ckl{\'e}, Andreas and Srivastava, Abhishek and Gurevych, Iryna},
  booktitle = {Thirty-fifth Conference on Neural Information Processing Systems, Datasets and Benchmarks Track},
  year      = {2021}
}

@inproceedings{muennighoff-etal-2023-mteb,
  title     = {{MTEB}: Massive Text Embedding Benchmark},
  author    = {Muennighoff, Niklas and Tazi, Nouamane and Magne, Lo{\"\i}c and Reimers, Nils},
  booktitle = {Proceedings of the 17th Conference of the European Chapter of the Association for Computational Linguistics (EACL)},
  year      = {2023}
}

@article{wang2022e5,
  title   = {Text Embeddings by Weakly-Supervised Contrastive Pre-training},
  author  = {Wang, Liang and Yang, Nan and Huang, Xiaolong and Jiao, Binxing and Yang, Linjun and Jiang, Daxin and Majumder, Rangan and Wei, Furu},
  journal = {arXiv preprint arXiv:2212.03533},
  year    = {2022}
}

@article{solatorio2023realtabformer,
  title={REaLTabFormer: Generating realistic relational and tabular data using transformers},
  author={Solatorio, Aivin V and Dupriez, Olivier},
  journal={arXiv preprint arXiv:2302.02041},
  year={2023}
}

@article{li2023gte,
  title   = {Towards General Text Embeddings with Multi-stage Contrastive Learning},
  author  = {Li, Zehan and Zhang, Xin and Zhang, Yanzhao and Long, Dingkun and Xie, Pengjun and Zhang, Meishan},
  journal = {arXiv preprint arXiv:2308.03281},
  year    = {2023}
}

@article{xiao2023bge,
  title   = {C-Pack: Packaged Resources To Advance General {C}hinese Embedding},
  author  = {Xiao, Shitao and Liu, Zheng and Zhang, Peitian and Muennighoff, Niklas},
  journal = {arXiv preprint arXiv:2309.07597},
  year    = {2023}
}

@article{solatorio2024gist,
  title   = {{GISTEmbed}: Guided In-sample Selection of Training Negatives for Text Embedding Fine-tuning},
  author  = {Solatorio, Aivin V.},
  journal = {arXiv preprint arXiv:2402.16829},
  year    = {2024}
}

@article{nogueira2019doc2query,
  title   = {Document Expansion by Query Prediction},
  author  = {Nogueira, Rodrigo and Yang, Wei and Lin, Jimmy and Cho, Kyunghyun},
  journal = {arXiv preprint arXiv:1904.08375},
  year    = {2019}
}

@inproceedings{bonifacio2022inpars,
  title     = {{InPars}: Unsupervised Dataset Generation for Information Retrieval},
  author    = {Bonifacio, Luiz and Abonizio, Hugo and Fadaee, Marzieh and Nogueira, Rodrigo},
  booktitle = {Proceedings of the 45th International ACM SIGIR Conference on Research and Development in Information Retrieval},
  year      = {2022}
}

@inproceedings{dai2023promptagator,
  title     = {Promptagator: Few-shot Dense Retrieval From 8 Examples},
  author    = {Dai, Zhuyun and Zhao, Vincent Y. and Ma, Ji and Luan, Yi and Ni, Jianmo and Lu, Jing and Bakalov, Anton and Guu, Kelvin and Hall, Keith B. and Chang, Ming-Wei},
  booktitle = {The Eleventh International Conference on Learning Representations (ICLR)},
  year      = {2023}
}

@inproceedings{wang-etal-2022-gpl,
  title     = {{GPL}: Generative Pseudo Labeling for Unsupervised Domain Adaptation of Dense Retrieval},
  author    = {Wang, Kexin and Thakur, Nandan and Reimers, Nils and Gurevych, Iryna},
  booktitle = {Proceedings of the 2022 Conference of the North American Chapter of the Association for Computational Linguistics (NAACL)},
  year      = {2022}
}

@inproceedings{faggioli2023llmjudge,
  title     = {Perspectives on Large Language Models for Relevance Judgment},
  author    = {Faggioli, Guglielmo and Dietz, Laura and Clarke, Charles L. A. and Demartini, Gianluca and Hagen, Matthias and Hauff, Claudia and Kando, Noriko and Kanoulas, Evangelos and Potthast, Martin and Stein, Benno and Wachsmuth, Henning},
  booktitle = {Proceedings of the 2023 ACM SIGIR International Conference on Theory of Information Retrieval (ICTIR)},
  year      = {2023}
}

@inproceedings{zheng2023judging,
  title     = {Judging {LLM}-as-a-Judge with {MT-Bench} and Chatbot Arena},
  author    = {Zheng, Lianmin and Chiang, Wei-Lin and Sheng, Ying and Zhuang, Siyuan and Wu, Zhanghao and Zhuang, Yonghao and Lin, Zi and Li, Zhuohan and Li, Dacheng and Xing, Eric P. and Zhang, Hao and Gonzalez, Joseph E. and Stoica, Ion},
  booktitle = {Thirty-seventh Conference on Neural Information Processing Systems, Datasets and Benchmarks Track},
  year      = {2023}
}

@article{henderson2017efficient,
  title   = {Efficient Natural Language Response Suggestion for Smart Reply},
  author  = {Henderson, Matthew and Al-Rfou, Rami and Strope, Brian and Sung, Yun-Hsuan and Luk{\'a}cs, L{\'a}szl{\'o} and Guo, Ruiqi and Kumar, Sanjiv and Miklos, Balint and Kurzweil, Ray},
  journal = {arXiv preprint arXiv:1705.00652},
  year    = {2017}
}

@inproceedings{karpukhin-etal-2020-dense,
  title     = {Dense Passage Retrieval for Open-Domain Question Answering},
  author    = {Karpukhin, Vladimir and O{\u{g}}uz, Barlas and Min, Sewon and Lewis, Patrick and Wu, Ledell and Edunov, Sergey and Chen, Danqi and Yih, Wen-tau},
  booktitle = {Proceedings of the 2020 Conference on Empirical Methods in Natural Language Processing (EMNLP)},
  year      = {2020}
}

@article{zhang2020webtable,
  title   = {Web Table Extraction, Retrieval, and Augmentation: A Survey},
  author  = {Zhang, Shuo and Balog, Krisztian},
  journal = {ACM Transactions on Intelligent Systems and Technology},
  volume  = {11},
  number  = {2},
  year    = {2020}
}

@inproceedings{yang-etal-2022-tableformer,
  title     = {{T}able{F}ormer: Robust Transformer Modeling for Table-Text Encoding},
  author    = {Yang, Jingfeng and Gupta, Aditya and Upadhyay, Shyam and He, Luheng and Goel, Rahul and Paul, Shachi},
  booktitle = {Proceedings of the 60th Annual Meeting of the Association for Computational Linguistics (ACL)},
  year      = {2022}
}

@inproceedings{lee-etal-2022-dedup,
  title     = {Deduplicating Training Data Makes Language Models Better},
  author    = {Lee, Katherine and Ippolito, Daphne and Nystrom, Andrew and Zhang, Chiyuan and Eck, Douglas and Callison-Burch, Chris and Carlini, Nicholas},
  booktitle = {Proceedings of the 60th Annual Meeting of the Association for Computational Linguistics (ACL)},
  year      = {2022}
}

@inproceedings{brickley2019datasetsearch,
  title     = {Google Dataset Search: Building a Search Engine for Datasets in an Open Web Ecosystem},
  author    = {Brickley, Dan and Burgess, Matthew and Noy, Natasha},
  booktitle = {Proceedings of The World Wide Web Conference (WWW)},
  year      = {2019}
}

@article{chapman2020dataset,
  title   = {Dataset Search: A Survey},
  author  = {Chapman, Adriane and Simperl, Elena and Koesten, Laura and Konstantinidis, George and Ib{\'a}{\~n}ez, Luis-Daniel and Kacprzak, Emilia and Groth, Paul},
  journal = {The VLDB Journal},
  volume  = {29},
  year    = {2020}
}

@inproceedings{thomas2024llmrel,
  title     = {Large Language Models Can Accurately Predict Searcher Preferences},
  author    = {Thomas, Paul and Spielman, Seth and Craswell, Nick and Mitra, Bhaskar},
  booktitle = {Proceedings of the 47th International ACM SIGIR Conference on Research and Development in Information Retrieval},
  year      = {2024}
}

@article{bajaj2016msmarco,
  title   = {{MS MARCO}: A Human Generated {MA}chine {R}eading {CO}mprehension Dataset},
  author  = {Bajaj, Payal and Campos, Daniel and Craswell, Nick and Deng, Li and Gao, Jianfeng and Liu, Xiaodong and Majumder, Rangan and McNamara, Andrew and Mitra, Bhaskar and Nguyen, Tri and Rosenberg, Mir and Song, Xia and Stoica, Alina and Tiwary, Saurabh and Wang, Tong},
  journal = {arXiv preprint arXiv:1611.09268},
  year    = {2016}
}

@article{izacard2022contriever,
  title   = {Unsupervised Dense Information Retrieval with Contrastive Learning},
  author  = {Izacard, Gautier and Caron, Mathilde and Hosseini, Lucas and Riedel, Sebastian and Bojanowski, Piotr and Joulin, Armand and Grave, Edouard},
  journal = {Transactions on Machine Learning Research},
  year    = {2022}
}

@book{tunkelang2009faceted,
  title     = {Faceted Search},
  author    = {Tunkelang, Daniel},
  series    = {Synthesis Lectures on Information Concepts, Retrieval, and Services},
  publisher = {Morgan \& Claypool},
  year      = {2009}
}

@inproceedings{robertson2004bm25f,
  title     = {Simple {BM25} Extension to Multiple Weighted Fields},
  author    = {Robertson, Stephen and Zaragoza, Hugo and Taylor, Michael},
  booktitle = {Proceedings of the 13th ACM International Conference on Information and Knowledge Management (CIKM)},
  year      = {2004}
}

@inproceedings{guo2009nerq,
  title     = {Named Entity Recognition in Query},
  author    = {Guo, Jiafeng and Xu, Gu and Cheng, Xueqi and Li, Hang},
  booktitle = {Proceedings of the 32nd International ACM SIGIR Conference on Research and Development in Information Retrieval},
  year      = {2009}
}

@inproceedings{cormack2009rrf,
  title     = {Reciprocal Rank Fusion Outperforms {C}ondorcet and Individual Rank Learning Methods},
  author    = {Cormack, Gordon V. and Clarke, Charles L. A. and Buettcher, Stefan},
  booktitle = {Proceedings of the 32nd International ACM SIGIR Conference on Research and Development in Information Retrieval},
  year      = {2009}
}

@inproceedings{conneau2020xlmr,
  title     = {Unsupervised Cross-lingual Representation Learning at Scale},
  author    = {Conneau, Alexis and Khandelwal, Kartikay and Goyal, Naman and Chaudhary, Vishrav and Wenzek, Guillaume and Guzm{\'a}n, Francisco and Grave, Edouard and Ott, Myle and Zettlemoyer, Luke and Stoyanov, Veselin},
  booktitle = {Proceedings of the 58th Annual Meeting of the Association for Computational Linguistics (ACL)},
  year      = {2020}
}

@article{wang2024me5,
  title   = {Multilingual E5 Text Embeddings: A Technical Report},
  author  = {Wang, Liang and Yang, Nan and Huang, Xiaolong and Yang, Linjun and Majumder, Rangan and Wei, Furu},
  journal = {arXiv preprint arXiv:2402.05672},
  year    = {2024}
}

@article{zhang2023miracl,
  title   = {{MIRACL}: A Multilingual Retrieval Dataset Covering 18 Diverse Languages},
  author  = {Zhang, Xinyu and Thakur, Nandan and Ogundepo, Odunayo and Kamalloo, Ehsan and Alfonso-Hermelo, David and Li, Xiaoguang and Liu, Qun and Rezagholizadeh, Mehdi and Lin, Jimmy},
  journal = {Transactions of the Association for Computational Linguistics},
  volume  = {11},
  year    = {2023}
}

@article{bonifacio2021mmarco,
  title   = {{mMARCO}: A Multilingual Version of the {MS MARCO} Passage Ranking Dataset},
  author  = {Bonifacio, Luiz and Jeronymo, Vitor and Abonizio, Hugo and Campiotti, Israel and Fadaee, Marzieh and Lotufo, Roberto and Nogueira, Rodrigo},
  journal = {arXiv preprint arXiv:2108.13897},
  year    = {2021}
}

@inproceedings{feng2022labse,
  title     = {Language-agnostic {BERT} Sentence Embedding},
  author    = {Feng, Fangxiaoyu and Yang, Yinfei and Cer, Daniel and Arivazhagan, Naveen and Wang, Wei},
  booktitle = {Proceedings of the 60th Annual Meeting of the Association for Computational Linguistics (ACL)},
  year      = {2022}
}

@inproceedings{qu2021rocketqa,
  title     = {{RocketQA}: An Optimized Training Approach to Dense Passage Retrieval for Open-Domain Question Answering},
  author    = {Qu, Yingqi and Ding, Yuchen and Liu, Jing and Liu, Kai and Ren, Ruiyang and Zhao, Wayne Xin and Dong, Daxiang and Wu, Hua and Wang, Haifeng},
  booktitle = {Proceedings of the 2021 Conference of the North American Chapter of the Association for Computational Linguistics (NAACL)},
  year      = {2021}
}

@inproceedings{xiong2021ance,
  title     = {Approximate Nearest Neighbor Negative Contrastive Learning for Dense Text Retrieval},
  author    = {Xiong, Lee and Xiong, Chenyan and Li, Ye and Tang, Kwok-Fung and Liu, Jialin and Bennett, Paul and Ahmed, Junaid and Overwijk, Arnold},
  booktitle = {International Conference on Learning Representations (ICLR)},
  year      = {2021}
}

@article{wilkinson2016fair,
  title   = {The {FAIR} Guiding Principles for Scientific Data Management and Stewardship},
  author  = {Wilkinson, Mark D. and Dumontier, Michel and Aalbersberg, IJsbrand Jan and Appleton, Gabrielle and others},
  journal = {Scientific Data},
  volume  = {3},
  number  = {160018},
  year    = {2016}
}

@article{schwartz2020green,
  title   = {Green {AI}},
  author  = {Schwartz, Roy and Dodge, Jesse and Smith, Noah A. and Etzioni, Oren},
  journal = {Communications of the ACM},
  volume  = {63},
  number  = {12},
  year    = {2020}
}

@inproceedings{strubell2019energy,
  title     = {Energy and Policy Considerations for Deep Learning in {NLP}},
  author    = {Strubell, Emma and Ganesh, Ananya and McCallum, Andrew},
  booktitle = {Proceedings of the 57th Annual Meeting of the Association for Computational Linguistics (ACL)},
  year      = {2019}
}

@inproceedings{lewis2020rag,
  title     = {Retrieval-Augmented Generation for Knowledge-Intensive {NLP} Tasks},
  author    = {Lewis, Patrick and Perez, Ethan and Piktus, Aleksandra and Petroni, Fabio and Karpukhin, Vladimir and Goyal, Naman and K{\"u}ttler, Heinrich and Lewis, Mike and Yih, Wen-tau and Rockt{\"a}schel, Tim and Riedel, Sebastian and Kiela, Douwe},
  booktitle = {Advances in Neural Information Processing Systems (NeurIPS)},
  year      = {2020}
}

@inproceedings{petrov2023tokenizers,
  title     = {Language Model Tokenizers Introduce Unfairness Between Languages},
  author    = {Petrov, Aleksandar and La Malfa, Emanuele and Torr, Philip H. S. and Bibi, Adel},
  booktitle = {Advances in Neural Information Processing Systems (NeurIPS)},
  year      = {2023}
}

@book{worldbank2021wdr,
  title     = {World Development Report 2021: Data for Better Lives},
  author    = {{World Bank}},
  publisher = {The World Bank},
  address   = {Washington, DC},
  year      = {2021}
}

@inproceedings{devlin2019bert,
  title     = {{BERT}: Pre-training of Deep Bidirectional Transformers for Language Understanding},
  author    = {Devlin, Jacob and Chang, Ming-Wei and Lee, Kenton and Toutanova, Kristina},
  booktitle = {Proceedings of the 2019 Conference of the North American Chapter of the Association for Computational Linguistics (NAACL)},
  year      = {2019}
}

@inproceedings{borisov2023great,
  title     = {Language Models are Realistic Tabular Data Generators},
  author    = {Borisov, Vadim and Se{\ss}ler, Kathrin and Leemann, Tobias and Pawelczyk, Martin and Kasneci, Gjergji},
  booktitle = {The Eleventh International Conference on Learning Representations (ICLR)},
  year      = {2023}
}

@inproceedings{li2023santa,
  title     = {Structure-Aware Language Model Pretraining Improves Dense Retrieval on Structured Data},
  author    = {Li, Xinze and Liu, Zhenghao and Xiong, Chenyan and Yu, Shi and Gu, Yu and Liu, Zhiyuan and Yu, Ge},
  booktitle = {Findings of the Association for Computational Linguistics: ACL 2023},
  year      = {2023}
}

@inproceedings{zaheer2017deepsets,
  title     = {Deep Sets},
  author    = {Zaheer, Manzil and Kottur, Satwik and Ravanbakhsh, Siamak and P{\'o}czos, Barnab{\'a}s and Salakhutdinov, Ruslan and Smola, Alexander J.},
  booktitle = {Advances in Neural Information Processing Systems (NeurIPS)},
  year      = {2017}
}

@inproceedings{murphy2019janossy,
  title     = {Janossy Pooling: Learning Deep Permutation-Invariant Functions for Variable-Size Inputs},
  author    = {Murphy, Ryan L. and Srinivasan, Balasubramaniam and Rao, Vinayak and Ribeiro, Bruno},
  booktitle = {International Conference on Learning Representations (ICLR)},
  year      = {2019}
}

@inproceedings{lee2019settransformer,
  title     = {Set Transformer: A Framework for Attention-based Permutation-Invariant Neural Networks},
  author    = {Lee, Juho and Lee, Yoonho and Kim, Jungtaek and Kosiorek, Adam R. and Choi, Seungjin and Teh, Yee Whye},
  booktitle = {International Conference on Machine Learning (ICML)},
  year      = {2019}
}

@inproceedings{chen2020grouptheoretic,
  title     = {A Group-Theoretic Framework for Data Augmentation},
  author    = {Chen, Shuxiao and Dobriban, Edgar and Lee, Jane H.},
  booktitle = {Advances in Neural Information Processing Systems (NeurIPS)},
  year      = {2020}
}

@article{liu2024lostmiddle,
  title   = {Lost in the Middle: How Language Models Use Long Contexts},
  author  = {Liu, Nelson F. and Lin, Kevin and Hewitt, John and Paranjape, Ashwin and Bevilacqua, Michele and Petroni, Fabio and Liang, Percy},
  journal = {Transactions of the Association for Computational Linguistics},
  volume  = {12},
  year    = {2024}
}

@inproceedings{lu2022fantastically,
  title     = {Fantastically Ordered Prompts and Where to Find Them: Overcoming Few-Shot Prompt Order Sensitivity},
  author    = {Lu, Yao and Bartolo, Max and Moore, Alastair and Riedel, Sebastian and Stenetorp, Pontus},
  booktitle = {Proceedings of the 60th Annual Meeting of the Association for Computational Linguistics (ACL)},
  year      = {2022}
}

@inproceedings{schein2002coldstart,
  title={Methods and Metrics for Cold-Start Recommendations},
  author={Schein, Andrew I. and Popescul, Alexandrin and Ungar, Lyle H. and Pennock, David M.},
  booktitle={Proceedings of the 25th Annual International ACM SIGIR Conference on Research and Development in Information Retrieval (SIGIR)},
  pages={253--260},
  year={2002}
}

@inproceedings{joachims2017unbiased,
  title={Unbiased Learning-to-Rank with Biased Feedback},
  author={Joachims, Thorsten and Swaminathan, Adith and Schnabel, Tobias},
  booktitle={Proceedings of the Tenth ACM International Conference on Web Search and Data Mining (WSDM)},
  pages={781--789},
  year={2017}
}

@inproceedings{park2008longtail,
  title={The Long Tail of Recommender Systems and How to Leverage It},
  author={Park, Yoon-Joo and Tuzhilin, Alexander},
  booktitle={Proceedings of the 2008 ACM Conference on Recommender Systems (RecSys)},
  pages={11--18},
  year={2008}
}
\bibliographystyle{iclr2026_conference}

\appendix

\section{Generation Prompts and Facet Definitions}
\label{app:prompts}
Training queries are generated by \texttt{claude-haiku-4-5} and evaluation queries
by \texttt{claude-sonnet-4-6}, under the identical prompt. The system prompt is:

\begin{quote}\scriptsize\begin{verbatim}
You are a development economist generating realistic search queries for a
development-data portal. You will receive the full metadata of ONE indicator.
Generate search queries for which this indicator is a correct retrieval result.

Rules:
- Each query is something a real researcher, policy analyst, journalist, or
  student would type.
- Follow the requested facet for each query (see facet definitions in the user
  message).
- For 'geo' / 'geo_year' facets: use a geography listed in the indicator's
  coverage (a country name, or a region that contains covered countries). Vary
  which geography you pick.
- For 'year' / 'geo_year' facets: use a year or range inside the indicator's
  time coverage.
- Vary phrasing, specificity, and persona. Do not copy the indicator name
  verbatim into more than one query. Mention the identifier code ONLY in
  'code'-facet queries.
- If a language is specified for the batch, write every query in that language
  (keep indicator codes, database acronyms, and proper nouns as-is).
- Return ONLY a JSON array: [{"facet": "...", "query": "..."}, ...]
\end{verbatim}\end{quote}

The user message supplies the full-schema serialization, the sampled facets, and
(for non-English) a language instruction, with these facet definitions:

\begin{quote}\scriptsize\begin{verbatim}
- keyword: 2-6 word keyword search
- natural: complete natural-language question
- definition: question about what a concept means / what the indicator measures
- methodology: question about how the data is collected, computed, or sourced
- geo: query that names a covered country or containing region
- year: query that names a year or period inside the time coverage
- geo_year: query combining a covered geography AND a covered year
- unit: query referencing the measurement unit or scale
- source: query referencing the source organization
- database: query referencing the database or collection the indicator belongs to
- frequency: query referencing the periodicity (e.g. annual, monthly, quarterly)
- thematic: broad policy/thematic framing related to the topics
\end{verbatim}\end{quote}

Under full coverage, one request asks for queries in five languages at once,
each tagged with a \texttt{lang} field, which sends the metadata once per group
and cuts input cost about fivefold. Identifier-lookup (\texttt{code}) queries are
templated rather than generated, from \texttt{\{c\}}, \texttt{indicator \{c\}},
\texttt{\{c\} series}, \texttt{\{c\} data}, \texttt{show indicator \{c\}}, and
\texttt{\{c\} dataset}, with \texttt{\{c\}} the identifier.

\section{Training Details and Hyperparameters}
\label{app:hparams}
All bases share the configuration in Table~\ref{tab:hparams}; only the base
encoder and (for the guided variants) the GIST guide vary. Training uses
sentence-transformers with AdamW. Serialization is performed on the fly in the
data loader, so every epoch presents fresh field permutations.

\begin{table}[h]
\caption{Shared training configuration.}
\label{tab:hparams}
\begin{center}\small
\begin{tabular}{ll}
\toprule
Setting & Value \\
\midrule
Loss & cached-MNRL (leaderboard); cached-GIST (guided) \\
Epochs & up to 5, early stopping (patience 5) \\
Batch size (negative pool) & 128 \\
GradCache mini-batch & 48 (halved on OOM) \\
Hard negatives per query & 3 \\
Learning rate & $3\times10^{-5}$, 10\% linear warmup \\
Max sequence length & 512 \\
Field-dropout probability & 0.15 (name and facet field protected) \\
Precision & bf16 \\
Eval / checkpoint interval & 1{,}500 steps \\
Early-stopping metric & loss on a held-back slice of training rows \\
GIST guide (default) & \texttt{harrier-oss-v1-0.6b}, fp16 inference \\
Serialization labels / budget & raw schema keys / generous \\
\bottomrule
\end{tabular}
\end{center}
\end{table}

Encoding conventions (for example the E5 \texttt{query:}/\texttt{passage:}
prefixes) are read from a model registry so training and evaluation cannot
diverge. The cached losses hold the negative pool (batch size) fixed and vary
only the GradCache mini-batch, so the contrastive objective is identical across
bases regardless of their memory footprint.

Wall-clock per fine-tune on a single GPU under the unguided cached-MNRL loss
(batch 128): \texttt{multilingual-e5-small} 12.2h, \texttt{harrier-270m} 24.8h,
\texttt{paraphrase-ml-minilm} 12.2h, \texttt{granite-97m} 9.9h,
\texttt{noinstruct-small} 10.5h, \texttt{gist-small} 10.4h, \texttt{minilm-l6}
5.3h; the no-permutation controls add a comparable amount each. The full
cross-base sweep of eleven fine-tunes totals 142 GPU-hours. GIST guiding with the
600M default guide multiplies a run's time several-fold, which the smaller
self-distilled guide largely avoids.

\section{Model Specifications and Selection Rationale}
\label{app:models}
Table~\ref{tab:models} lists every model. Trainees span multilingual encoders
and, as a controlled study of the vocabulary ceiling (Section~\ref{sec:results-tokenizer}),
English-vocabulary encoders. \texttt{harrier-oss-v1-0.6b} serves as both the
hard-negative miner and the default GIST guide because it is the strongest open
multilingual encoder we had; the OpenAI models are frontier zero-shot references.

\begin{table}[h]
\caption{Models, with parameter count, vocabulary type, role, and query/document
prefixes. ft = fine-tuned trainee; zs = zero-shot baseline.}
\label{tab:models}
\begin{center}\small
\begin{tabular}{lllll}
\toprule
Model & Params & Vocabulary & Role & Prefixes \\
\midrule
multilingual-e5-small & 118M & multilingual & ft & query:/passage: \\
paraphrase-ml-minilm & 118M & multilingual & ft & none \\
harrier-oss-v1-270m & 270M & multilingual & ft & none \\
granite-embedding-97m-r2 & 97M & multilingual & ft & none \\
noinstruct-small & 33M & English & ft & none \\
gist-small & 33M & English & ft & none \\
all-MiniLM-L6-v2 & 22M & English & ft & none \\
\midrule
harrier-oss-v1-0.6b & 600M & multilingual & miner, guide, zs & none \\
jina-embeddings-v5-nano & 38M & multilingual & zs & task prompts \\
bge-small-en-v1.5 & 33M & English & zs & query instruction \\
gte-small & 33M & English & zs & none \\
text-embedding-3-small & API & multilingual & zs & none \\
text-embedding-3-large & API & multilingual & zs & none \\
\bottomrule
\end{tabular}
\end{center}
\end{table}

NoInstruct is trained in two forms, symmetric mean pooling and the model card's
prescribed asymmetric pooling (mean queries, CLS documents); the symmetric form
is slightly better once fine-tuned (Section~\ref{sec:results-ablations}).

\section{Extended Audits}
\label{app:audits}

\subsection{Full robustness across all models}
\label{app:robustfull}
\begin{figure}[h]
\centering
\includegraphics[width=0.96\linewidth]{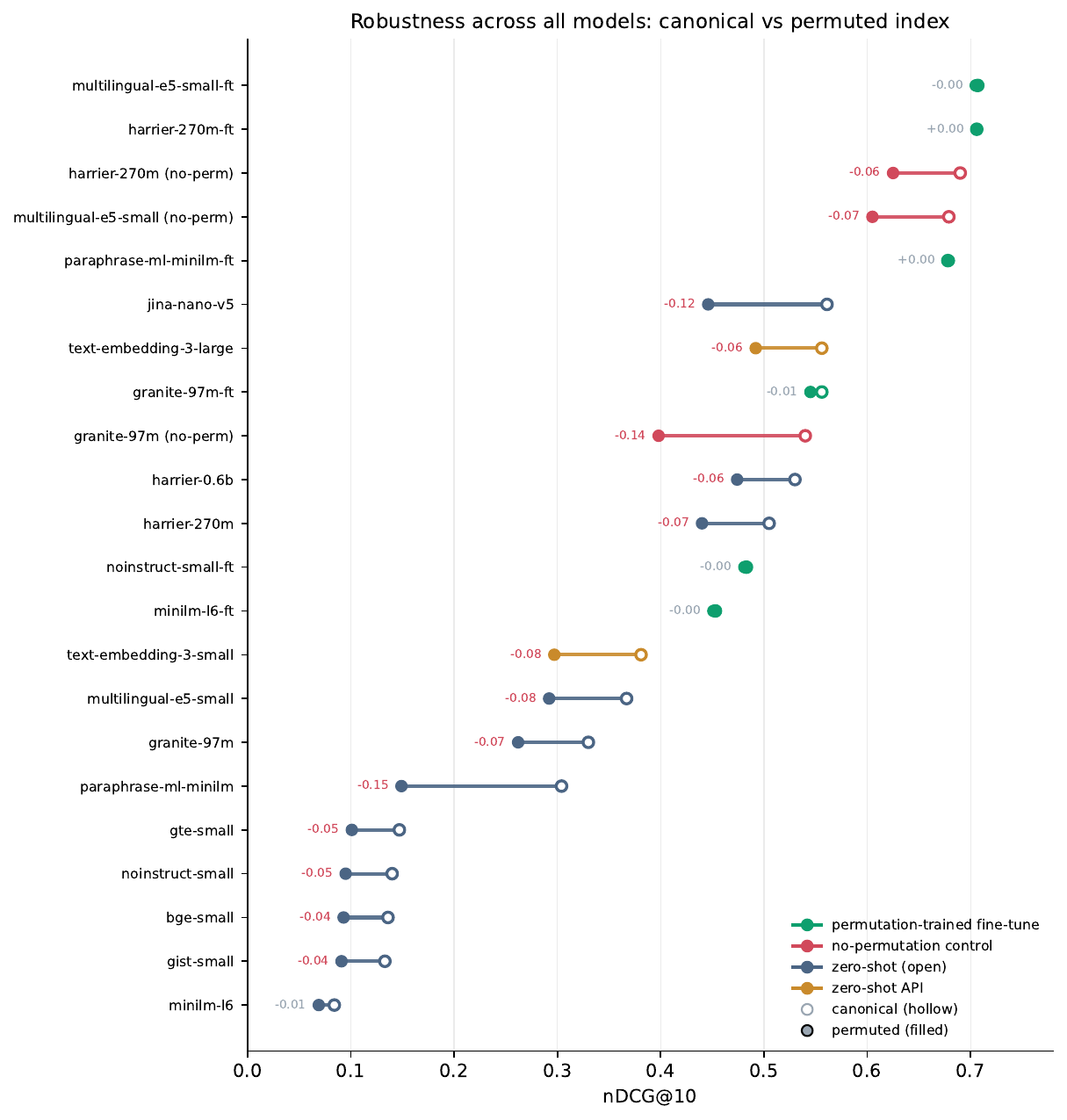}
\caption{Robustness to a serialization-order change across all 22 systems:
canonical (hollow) to permuted (filled) nDCG@10, colored by category, with the
order-change delta. Permutation-trained fine-tunes are invariant; no-permutation
controls, zero-shot open models, and API embeddings all degrade.}
\label{fig:robustfull}
\end{figure}

Figure~\ref{fig:robustfull} gives the canonical-versus-permuted index comparison
of Section~\ref{sec:results-robustness} for every system: the permutation-trained
fine-tunes, the no-permutation controls, all zero-shot open baselines, and the
two API embeddings. Each model is a dumbbell from its canonical nDCG@10 (hollow)
to its permuted nDCG@10 (filled); the number is the change. Only the
permutation-trained fine-tunes are flat. Every other system, regardless of its
absolute quality, loses ground under the order change.

\subsection{Per-topic retrieval}
\label{app:topics}
Topics are the catalog's own subject taxonomy, and each indicator carries about
2.4 of them, so a query inherits the topics of its target. Table~\ref{tab:topic}
reports R@10 over the twelve highest-volume topics among the held-out positives.
Two patterns recur across models. The finance and macro-policy topics
(Macro-financial Policies, Economic Policy, Financial Stability) are the hardest,
which tracks the near-duplicate density of those series, where many agencies
republish overlapping indicators. The institution and trade topics are the
easiest, scoring high even for the zero-shot base. Fine-tuning lifts the hard
finance topics most: the e5 base rises from 0.250 to 0.805 on Macro-financial
Policies and from 0.316 to 0.836 on Economic Policy, so the framework helps most
exactly where the corpus is most saturated with near-twins.

\begin{table}[t]
\caption{Per-topic R@10 on the multilingual holdout, the twelve most frequent
topics among held-out positives. T1 Prosperity, T2 Economic Policy,
T3 Macro-financial Policies, T4 Finance, T5 Financial Stability and Integrity,
T6 Trade, Investment and Competitiveness, T7 People, T8 Institutions,
T9 Fiscal Policy, T10 Investment and Business Climate, T11 Public Institutions,
T12 Education. A positive may carry several topics, so the columns are not a
partition.}
\label{tab:topic}
\begin{center}\footnotesize
\setlength{\tabcolsep}{4pt}
\begin{tabular}{lcccccccccccc}
\toprule
Model & T1 & T2 & T3 & T4 & T5 & T6 & T7 & T8 & T9 & T10 & T11 & T12 \\
\midrule
multilingual-e5-small (0-shot) & 0.451 & 0.316 & 0.250 & 0.360 & 0.300 & 0.697 & 0.609 & 0.715 & 0.421 & 0.712 & 0.745 & 0.557 \\
\midrule
multilingual-e5-small-ft & \textbf{0.869} & \textbf{0.836} & \textbf{0.805} & \textbf{0.876} & \textbf{0.859} & 0.939 & 0.901 & 0.915 & 0.878 & \textbf{0.954} & 0.931 & \textbf{0.884} \\
harrier-270m-ft & 0.852 & 0.806 & 0.760 & 0.854 & 0.839 & \textbf{0.940} & \textbf{0.920} & \textbf{0.923} & \textbf{0.881} & 0.947 & \textbf{0.933} & \textbf{0.884} \\
granite-97m-ft & 0.715 & 0.636 & 0.570 & 0.710 & 0.676 & 0.856 & 0.803 & 0.827 & 0.746 & 0.869 & 0.818 & 0.773 \\
noinstruct-small-ft & 0.669 & 0.644 & 0.612 & 0.658 & 0.667 & 0.749 & 0.680 & 0.739 & 0.685 & 0.773 & 0.769 & 0.703 \\
\bottomrule
\end{tabular}
\end{center}
\end{table}

\subsection{Robustness of the near-duplicate conditioning}
\label{app:neardup}
The conditioning result in Section~\ref{sec:results-neardup} does not depend on
how a duplicate is defined. Table~\ref{tab:neardup-robust} reports the
distinct-minus-duplicate R@10 gap under three definitions: the primary
NoInstruct-small embedding (84\% of held-out indicators flagged), a stronger
multilingual-e5-small embedding (96\%), and an embedding-independent lexical
criterion (token Jaccard $\geq 0.7$ on name and definition, 81\%). The 0.95
cutoff is a single threshold on a continuous quantity; the median
nearest-neighbor similarity is 0.977 and 42.9\% of indicators exceed 0.98.
Across all three definitions the multilingual fine-tunes are non-negative, so
they do at least as well on the distinct records; the zero-shot base shows a
large positive gap; and the English-vocabulary fine-tune shows a small negative
gap. The e5 embedding flags 96\% of records as duplicates, leaving only 4\%
distinct, and the fine-tunes still score higher on that small distinct set,
which is the strongest form of the argument.

\begin{table}[t]
\caption{Distinct-minus-duplicate R@10 gap ($\Delta$) under three duplicate
definitions (multilingual holdout). Positive means higher R@10 on records with
no close training neighbor. Parenthetical percentages are the share of held-out
indicators flagged as near-duplicates by each definition.}
\label{tab:neardup-robust}
\begin{center}\small
\begin{tabular}{lccc}
\toprule
Model & NoInstruct emb (84\%) & e5 emb (96\%) & lexical (81\%) \\
\midrule
multilingual-e5-small-ft & $+0.025$ & $+0.030$ & $+0.002$ \\
harrier-270m-ft & $+0.062$ & $+0.056$ & $+0.034$ \\
granite-97m-ft & $+0.074$ & $-0.001$ & $+0.031$ \\
multilingual-e5-small (0-shot) & $+0.216$ & $+0.122$ & $+0.136$ \\
noinstruct-small-ft (Eng-vocab) & $-0.013$ & $-0.045$ & $-0.038$ \\
\bottomrule
\end{tabular}
\end{center}
\end{table}

The near-duplicate similarity distribution is shown in Figure~\ref{fig:corpus}c. Further audits, including the full per-language tokenizer table, the geographic coverage of evaluation queries, and per-economy variance, are deferred to an extended version.

\section{Cost Accounting}
\label{app:cost}
Generation used a small model (\texttt{claude-haiku-4-5} for training,
\texttt{claude-sonnet-4-6} for evaluation) through a 50\%-discounted batch API,
and the identifier-lookup queries are templated rather than generated, so the
benchmark is inexpensive to regenerate from the public catalog.

\end{document}